\documentclass[10pt]{article}
\PassOptionsToPackage{table}{xcolor}
\usepackage[utf8]{inputenc}
\usepackage[accepted]{tmlr}

\usepackage{amsmath,amsfonts,bm}

\def\eqref#1{equation~\ref{#1}}
\def\Eqref#1{Equation~\ref{#1}}

\def\1{\bm{1}}

\DeclareMathAlphabet{\mathsfit}{\encodingdefault}{\sfdefault}{m}{sl}
\SetMathAlphabet{\mathsfit}{bold}{\encodingdefault}{\sfdefault}{bx}{n}

\usepackage{graphicx}
\usepackage{float}
\usepackage{booktabs}
\usepackage{algorithm}
\usepackage{multirow}
\usepackage{xcolor}
\usepackage{longtable}
\usepackage{pdflscape}
\usepackage{mathtools}
\usepackage{tabularx}
\usepackage{array}
\usepackage{adjustbox}
\usepackage{ragged2e}
\usepackage{amsthm}
\usepackage[hidelinks]{hyperref}
\usepackage[capitalize,nameinlink]{cleveref}
\usepackage{url}
\usepackage{makecell}
\usepackage{siunitx}
\usepackage{comment}

\definecolor{boxgray}{gray}{0.95}
\definecolor{bestgreen}{RGB}{180,230,180}
\definecolor{secondgreen}{RGB}{220,245,220}
\definecolor{worstred}{RGB}{245,180,180}
\definecolor{secondred}{RGB}{250,215,215}
\definecolor{tirow}{RGB}{239,247,255}
\definecolor{poolrow}{RGB}{247,246,242}
\definecolor{dprow}{RGB}{242,249,243}

\newcommand{\cmark}{\ensuremath{\surd}}
\newcommand{\xmark}{\ensuremath{\times}}
\providecommand{\NA}{\multicolumn{1}{c}{---}}
\providecommand{\ninv}{\textsuperscript{\(\times\)}}

\newcommand{\appref}[1]{Appendix~\ref{#1}}

\usepackage{soul}
\usepackage{etoolbox}
\definecolor{revred}{HTML}{C1121F}
\newif\ifshowrevisions
\showrevisionsfalse
\ifshowrevisions
  \newcommand{\rev}[1]{\textcolor{revred}{#1}}
  \newcommand{\revdel}[1]{\textcolor{revred}{\st{#1}}}
  \newcommand{\revnewcaption}[1]{\color{revred}\caption{#1}\color{black}}
  \newcommand{\revnewsection}[1]{\color{revred}\section{#1}\color{black}}
  
  \newcommand{\sig}{\textsuperscript{\textcolor{revred}{$\star$}}}
  \AtBeginEnvironment{thebibliography}{%
    \let\revoldbibitem\bibitem
    \renewcommand{\bibitem}[2][]{%
      \color{black}%
      \ifstrequal{#2}{xie2021segformer}{\color{revred}}{}%
      \revoldbibitem[#1]{#2}}}
\else
  \newcommand{\rev}[1]{#1}
  \newcommand{\revdel}[1]{}
  \newcommand{\revnewcaption}[1]{\caption{#1}}
  \newcommand{\revnewsection}[1]{\section{#1}}
  
  \newcommand{\sig}{\textsuperscript{$\star$}}
\fi

\newcolumntype{P}[1]{>{\RaggedRight\arraybackslash}p{#1}}

\renewcommand{\arraystretch}{1.05}
\title{Rethinking Post-Hoc Calibration in Semantic Segmentation}

\author{%
\name Tristan Kirscher$^{1,2}$$^{\star}$,\;
Kim-Celine Kahl$^{3,4}$,\;
Balint Kovacs$^{3,5}$,\;
Maximilian Rokuss$^{3,4}$,\;
Klaus Maier-Hein$^{3,6}$,\;
Xavier Coubez$^{1,2}$,\;
Philippe Meyer$^{1,2}$,\;
Sylvain Faisan$^{1}$ \\
\addr
$^{1}$ ICube Laboratory, CNRS UMR 7357, University of Strasbourg, Strasbourg, France \\
$^{2}$ CLCC Institut Strauss, Strasbourg, France \\
$^{3}$ German Cancer Research Center (DKFZ), Division of Medical Image Computing, Heidelberg, Germany \\
$^{4}$ Faculty of Mathematics and Computer Science, University of Heidelberg, Germany \\
$^{5}$ Medical Faculty Heidelberg, Heidelberg University, Heidelberg, Germany \\
$^{6}$ Pattern Analysis and Learning Group, Dept.\ of Radiation Oncology, Heidelberg University Hospital, Germany \\[0.6em]
$^{\star}$ Corresponding author: \texttt{tristan.kirscher@unistra.fr}.%
}

\def\month{09}
\def\year{2026}
\def\openreview{\url{https://openreview.net/forum?id=xwNoSNxgxV}}

\begin{document}
\maketitle

\begin{abstract}
Reliable confidence estimates are essential in semantic segmentation, especially in safety-critical settings where overconfident errors can mislead downstream decisions. Yet modern segmentation models often remain miscalibrated.
Post-hoc calibration offers a practical way to correct confidence estimates without retraining the segmentation model, but its use in dense prediction raises structural issues that are often overlooked.
We study two such issues.
First, adding a constant to all logits leaves the softmax probabilities unchanged, but several standard calibrators can still depend on this arbitrary offset. As a result, two logit representations encoding the same predictive distribution may yield different calibrated probabilities. \rev{This non-identifiability also exists in image classification. In segmentation, however, the offset can vary across pixels or voxels, creating spatially varying representation dependence.} We define translation-invariant (TI) calibrators as those whose outputs are unchanged under such shifts, characterize which common calibrators satisfy this property, and construct TI counterparts of shift-sensitive calibrators to isolate the effect of removing representation dependence.
Second, post-hoc calibration is typically fitted by minimizing a likelihood-based objective, whereas segmentation models are trained with \rev{objectives that combine cross-entropy with task-specific terms such as soft Dice}\revdel{task-specific metrics such as Dice}. \revdel{This mismatch can cause calibration to alter class orderings and degrade the deployed segmentation map.} \rev{Fitting a calibrator on a held-out set using cross-entropy loss can degrade the quality of the segmentation map, due to the mismatch between the training and calibration losses and the small size of the calibration set.}
We study decision-preserving calibration under argmax- and order-preservation constraints. Since enforcing these constraints collapses affine softmax calibrators to temperature scaling, we introduce class-conditional affine calibrators that can be made argmax- or order-preserving while retaining greater expressivity, allowing us to quantify the calibration--segmentation trade-off induced by decision preservation.
Across natural-image and medical segmentation benchmarks, and under corruption-based covariate shift, matched comparisons show that TI variants generally improve calibration metrics, while decision-preserving variants prevent segmentation degradation \rev{by construction} and retain strong calibration performance.
These results provide practical design principles for well-defined post-hoc calibration pipelines in semantic segmentation.
\end{abstract}

\begin{figure}[t]
    \centering
    \includegraphics[width=1\linewidth]{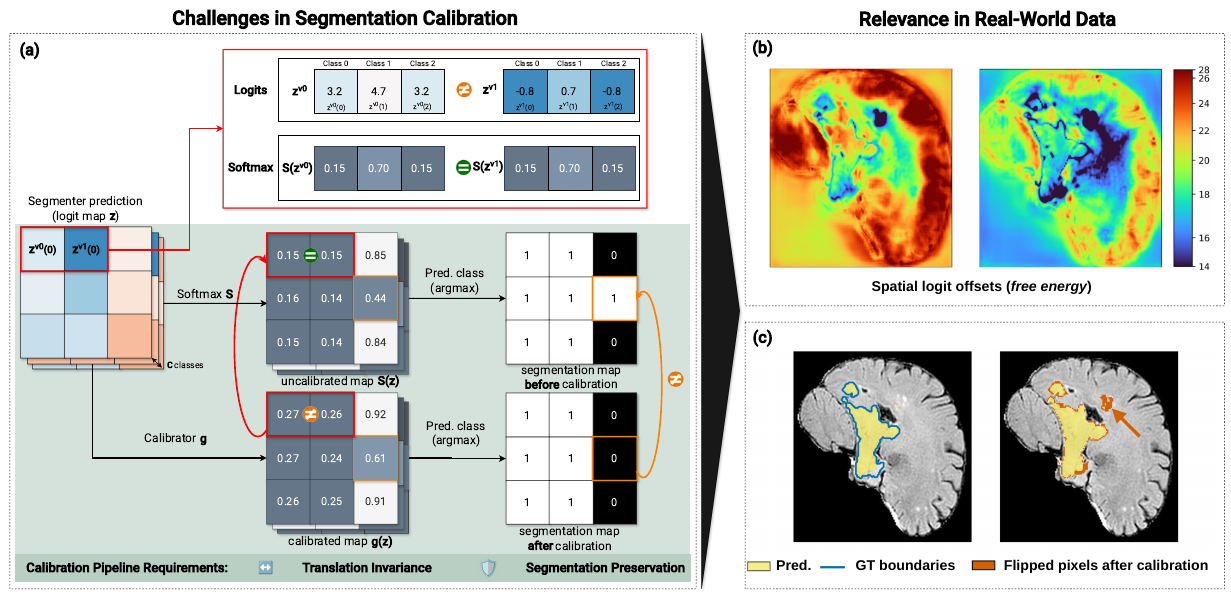}
    \caption{\textbf{Challenges in Segmentation Calibration: Evidence from Real-World Data.}
    \textbf{(1) Translation invariance.} As illustrated in panel (a), logit embeddings that differ only by a constant offset induce identical class probabilities under softmax, yet a calibrator may produce different outputs for these equivalent representations. Panel (b) shows, for two segmenters, spatial maps of the logit free energy, highlighting that this additive logit component varies across locations in real segmentation outputs. This motivates calibration pipelines that are invariant to such offsets.
    \textbf{(2) Segmentation preservation.} As shown in panel (a), calibration may change the argmax, flipping labels and altering the segmentation. Panel (c) shows a real example where such flips perturb brain tumor boundaries. A desirable post-hoc calibrator therefore preserves the deployed segmentation while improving confidence estimates.
}
    \label{fig:overview}
\end{figure}

\section{Introduction}
\label{sec:intro}
Semantic segmentation models can achieve strong task performance while still producing unreliable confidence estimates. In dense prediction settings, calibration is particularly important because voxel- or pixel-wise confidence scores are used for downstream decisions, quality control, uncertainty-aware post-processing, and failure detection~\citep{mehrtash_confidence_2020,kahl2024values,zenk2025benchmarking}. However, high segmentation accuracy does not guarantee reliable predictive probabilities: segmentation models may be miscalibrated, exhibiting over- or underconfidence, especially near object boundaries, ambiguous structures, or distribution shifts~\citep{rahaman_uncertainty_2021,zeevi_spatially-aware_2025}. Such errors can undermine the practical utility of confidence estimates even when the predicted segmentation masks are accurate.


Post-hoc calibration offers a practical way to correct confidence estimates after training, without retraining or modifying the underlying segmentation model. Train-time calibration methods, such as label smoothing~\citep{muller2019when}, differentiable calibration losses~\citep{kumar2018trainable}, and focal-loss-based calibration~\citep{mukhoti2020calibrating}, are complementary. We focus on post-hoc calibration because it can be applied to already deployed segmentation models without retraining, can be updated under changing deployment conditions, including distribution shifts, and allows confidence correction to be studied separately from segmentation-model training.  Standard post-hoc methods typically learn a global transformation in logit or probability space on a held-out calibration set, often by minimizing cross-entropy~\citep{guo_calibration_2017,kull2019beyond,zhang2020mix}. These methods are largely inherited from image-level classification, where each sample is associated with a single prediction. In semantic segmentation, however, calibration is applied densely across many spatial locations, and the interaction between logit-space transformations, spatially structured predictions, and task-specific segmentation metrics has not been fully characterized.

A fundamental but often overlooked point is that logits are non-identifiable: adding a constant to all class logits at a given spatial location leaves the softmax probabilities unchanged. Thus, a predictive distribution corresponds not to a unique logit vector, but to an equivalence class of logits that differ by additive shifts, sometimes interpreted as a ``gauge'' degree of freedom. Consequently, two logit vectors that induce the same predictive distribution can yield different calibrated outputs if the calibration function operates directly in logit space and is sensitive to such shifts. This dependence on an arbitrary logit representative introduces representation sensitivity, potentially leading to inconsistent or unstable calibration behavior.

While logit non-identifiability also holds in image-level classification, its practical impact may be amplified in semantic segmentation. Calibration is applied across many voxels or pixels, and additive offsets may vary across spatial locations, effectively introducing a field of local gauge degrees of freedom.
This spatial structure increases the dimensionality of the ambiguity and may make its practical consequences more pronounced for dense prediction tasks. \rev{Translation invariance is therefore a general property of multiclass calibration; what dense prediction changes is that the offset varies within a single prediction.}
Fig.~\ref{fig:overview}(a, b) illustrates this issue through shifted-logit examples and empirical free-energy maps, showing that such offsets occur in practice and vary spatially.

A second structural issue arises from the mismatch between training and calibration objectives. Segmentation models are typically trained with task-specific losses such as the Dice loss or its variants~\citep{sudre2017generalised,isensee_nnu-net_2021}\rev{, in practice usually inside a compound objective that also contains cross-entropy}, whereas post-hoc calibration usually minimizes cross-entropy~\citep{guo_calibration_2017,kull2019beyond}. Because optimizing cross-entropy can alter class orderings, calibration may change the predicted segmentation mask and potentially reduce the DSC.
In our experiments, such calibration-induced decision changes degrade segmentation performance (Fig.~\ref{fig:overview}(c)), motivating calibrators that correct confidence estimates without altering the original decision map. \rev{That mismatch is not the only driver of this cost. Post-hoc calibration is by construction fitted on a small held-out calibration set, and expressive calibration maps are known to overfit it and to lose accuracy even in plain classification, which is what motivates regularized and accuracy-preserving forms~\citep{guo_calibration_2017,kull2019beyond,zhang2020mix,rahimi2020}.}

Despite the widespread use of post-hoc calibration in segmentation, the interplay between logit non-identifiability, spatially dense calibration, and calibration-induced decision changes has not been systematically analyzed. We address this gap through two complementary structural constraints on post-hoc calibration:

\begin{itemize}
\item We formalize \emph{translation invariance} as a condition for representation-consistent calibration, characterize which common calibrators satisfy it, and construct matched translation-invariant (TI) counterparts of shift-sensitive calibrators to isolate the effect of removing representation dependence.
\item We study decision-preserving calibration under argmax- and order-preservation constraints. Since enforcing these constraints collapses affine softmax calibrators to temperature scaling, we introduce class-conditional affine calibrators that can be made argmax- or order-preserving while retaining greater expressivity, allowing us to quantify the calibration--segmentation trade-off induced by decision preservation.
\item Across natural-image and medical segmentation benchmarks, and under corruption-based covariate shift, paired comparisons show that TI variants improve calibration metrics in most matched comparisons. Decision-preserving variants prevent calibration-induced DSC drops by construction, while retaining strong calibration performance.
\end{itemize}

Overall, our goal is not to identify a universally optimal calibration pipeline, since performance also depends on calibrator expressivity and dataset-specific characteristics. Instead, we clarify which structural properties make post-hoc calibration well-defined for dense prediction. Translation invariance ensures that calibration does not depend on an arbitrary logit representative, while argmax or order preservation prevents calibration from altering the deployed segmentation map.

\section{Related Work}
\label{sec:related}

\paragraph{Post-hoc calibration of neural networks.}
Post-hoc calibration aims to align predicted confidence with empirical accuracy without modifying the trained model. Pioneering work focused on binary or small-scale classification, including Platt scaling~\citep{platt1999probabilistic}, histogram binning~\citep{zadrozny2001}, isotonic regression~\citep{zadrozny2002}, and Bayesian binning into quantiles~\citep{naeini2015}. For modern multiclass neural networks, commonly used methods include temperature scaling (TS), vector scaling (VS), matrix scaling (MS), and Dirichlet calibration (DC)~\citep{guo_calibration_2017,kull2019beyond}, as well as order-preserving and Gaussian process-based extensions~\citep{rahimi2020,milios2018,wenger2020}.
A fundamental but often overlooked issue in calibration is that logits are not uniquely defined: adding a constant to all logits leaves softmax probabilities unchanged. The implications of this invariance for post-hoc calibration have received little attention. In practice, calibration is often performed directly in logit space using affine transformations such as vector or matrix scaling, which may depend on the particular logit representative chosen. As a result, calibrated predictions may vary across equivalent logit embeddings. Although one could argue that logit shifts may encode useful information in certain contexts, for example in energy-based formulations for out-of-distribution detection~\citep{liu2020energy,kim2024uncertainty}, post-hoc calibration of predictive probabilities should not depend on arbitrary representational degrees of freedom unless explicitly modeled. In contrast to prior work, we formalize translation invariance as a necessary condition for representation-consistent calibration and characterize which calibration pipelines satisfy it, or how this property can be enforced.

\paragraph{Post-hoc calibration in semantic segmentation.}
In semantic segmentation, reliable voxel- or pixel-wise confidence estimates are important for downstream decisions, quality control, failure detection, and uncertainty-aware post-processing~\citep{mehrtash_confidence_2020,kahl2024values,zenk2025benchmarking}. Calibration methods developed for classification are commonly transferred to dense prediction by treating pixels or voxels as independent calibration samples and applying a global mapping~\citep{rousseau_post_2025}. This global approach assumes uniform calibration across the image, although calibration errors often vary spatially, especially near object boundaries and ambiguous structures~\citep{zeevi_spatially-aware_2025,liang_we_2025}. To address spatial heterogeneity, adaptive methods such as Local Temperature Scaling (LTS)~\citep{ding2021localtemperaturescaling} learn spatially varying temperature maps, where the temperature at each location is predicted from the network's outputs, such as logits. Selective scaling mechanisms have also been proposed to enable more flexible calibration in semantic segmentation by focusing adjustment on misclassified pixels~\citep{wang2023}.
A second challenge is the mismatch between calibration and task-specific segmentation objectives. Segmentation models are often trained or evaluated with overlap-based criteria such as Dice~\citep{sudre2017generalised,isensee_nnu-net_2021}\rev{, in practice usually combined with cross-entropy}, whereas post-hoc calibration is typically fitted by minimizing cross-entropy. Consequently, a calibrator can improve the calibration loss while changing class orderings, thereby modifying the predicted segmentation map and causing measurable DSC drops~\citep{rousseau_post_2025}. This motivates decision-preserving calibration, in which calibrated probabilities preserve either the original argmax or the complete class ordering. In segmentation, methods with this property commonly rely on temperature-scaling variants~\citep{ding2021localtemperaturescaling,wang2023}, which are naturally order-preserving.
Decision and order preservation have also been considered in classification. Mix-n-Match~\citep{zhang2020mix} promotes accuracy-preserving calibration and introduces ETS, a convex mixture of temperature scaling, the identity map, and the uniform distribution, to increase expressivity while preserving accuracy. \cite{rahimi2020} go further by learning intra order-preserving calibration maps with a constrained neural architecture that preserves top-$k$ predictions. \rev{For classification, the loss of accuracy that can be observed after calibration cannot be attributed to objective mismatch but to the post-hoc setting itself: expressive maps such as matrix scaling and Dirichlet calibration are fitted on a small held-out set and may lose accuracy due to overfitting, which motivates regularized and accuracy-preserving approaches~\citep{guo_calibration_2017,kull2019beyond,zhang2020mix}.}
In dense prediction, \revdel{this issue} \rev{decision and order preservation} has received comparatively less attention, although the loss mismatch discussed above makes it particularly important.
We propose to construct matched decision-preserving counterparts of expressive unconstrained calibrators, in order to isolate the effect of this constraint and quantify the calibration--segmentation trade-off. The main difficulty is to impose argmax or order preservation without reducing the calibrator to temperature scaling. To this end, we introduce class-conditional affine calibrators that can be made to preserve either the original argmax or the complete class ordering while retaining greater expressivity.

\section{Methods}
\label{sec:methods}
\subsection{Translation-Invariant Post-hoc Calibration}
\label{sec:translation}
For a $C$-class segmentation task, each location $v$ is associated with a ground-truth label $y^v \in \{1,\dots,C\}$ and a logit vector $\mathbf z^v \in\mathbb R^C$ produced by a segmentation model. Class probabilities are obtained through the softmax transformation $\mathbf p^v=S(\mathbf z^v)$. Post-hoc calibration learns a mapping $g$ such that the calibrated probabilities $\mathbf q^v = g(\mathbf z^v)$ better reflect empirical accuracies. In the case of top-label calibration, this can be expressed as
\begin{equation}
\label{eq:calib}
\mathbb P\!\left(y^v=\hat y \mid \mathbf q^v_{\hat y}\right)
=
\mathbf q^v_{\hat y},
\end{equation}
where $\hat y = \arg\max_c \mathbf q^v_c$. In practice, $g$ is fitted by minimizing a proper scoring rule, typically the negative log-likelihood (NLL), on a calibration set.

Since the softmax is invariant to additive shifts, two logit vectors $\mathbf{z}^{v_1}$ and $\mathbf{z}^{v_2}$ differing by a common additive constant yield identical probability distributions ($S(\mathbf{z}^{v_1}) = S(\mathbf{z}^{v_2})$), but 
applying a calibrator in logit space can break this equivalence, producing different calibrated predictions: 
$g(\mathbf{z}^{v_1}) \neq g(\mathbf{z}^{v_2})$. 
To preserve this equivalence, we define a calibrator as translation-invariant (TI) if it satisfies
\begin{equation}
\label{eq:translation_inv}
g(\mathbf{z}) = g(\mathbf{z} + c\mathbf{1}),
\quad \forall \mathbf{z} \in \mathbb{R}^C,\; \forall c \in \mathbb{R}.
\end{equation}

Under this property, any two logit vectors yielding the same softmax
probabilities necessarily produce identical calibrated predictions.
Equivalently, the output of a TI calibrator depends only on the predictive
distribution induced by the softmax and not on the particular logit
representation.
Translation invariance is a natural consistency requirement for post-hoc calibration of predictive probabilities. Otherwise, the calibrator would depend on a degree of freedom that carries no predictive information.

\paragraph{Standard post-hoc calibrators.}
We consider standard multiclass post-hoc calibrators applied voxel-wise, including temperature scaling (TS) and its affine extensions, vector scaling (VS) and matrix scaling (MS)~\citep{guo_calibration_2017}. We further study two TS-based variants: ensemble temperature scaling (ETS)~\citep{zhang2020mix}, which mixes the temperature-scaled distribution with the original softmax output and a uniform prior, and local temperature scaling (LTS)~\citep{ding2021localtemperaturescaling}, where a temperature function $\tau_\theta$ (e.g., a CNN) predicts a spatially varying temperature field from the full logit map $\mathbf{Z}$ and optional features $\xi$:

\begin{align}
g^{\mathrm{TS}}_{T}(\mathbf{z})
&= S(\mathbf{z}/T), && T>0,
\label{eq:calib-ts} \\
g^{\mathrm{VS}}_{\mathbf a,\mathbf b}(\mathbf{z})
&= S(\operatorname{diag}(\mathbf{a})\,\mathbf{z} + \mathbf{b}),
&& \mathbf{a},\mathbf{b}\in\mathbb{R}^C,
\label{eq:calib-vs} \\
g^{\mathrm{MS}}_{\mathbf W,\mathbf b}(\mathbf{z})
&= S(\mathbf{W}\mathbf{z} + \mathbf{b}),
&& \mathbf{W}\in\mathbb{R}^{C\times C},\; \mathbf{b}\in\mathbb{R}^C,
\label{eq:calib-ms} \\
g^{\mathrm{ETS}}_{\lambda,T}(\mathbf z)
&=
\lambda_1 S(\mathbf z/T)
+
\lambda_2 S(\mathbf z)
+
\lambda_3 \mathbf u,
&&
\lambda_k\ge 0,\;
\sum_{k=1}^3\lambda_k=1,\;
\mathbf u=\frac{1}{C}\mathbf 1,
\label{eq:calib-ets} \\
g_{\theta}^{\mathrm{LTS}}(\mathbf z^v)
&=
S\!\left(
\mathbf z^v/\tau_\theta(v;\mathbf Z,\xi)
\right),
&& \tau_\theta(v;\mathbf Z,\xi)>0.
\label{eq:calib-lts}
\end{align}

The following properties follow directly from the definitions. TS and ETS are translation-invariant (TI). VS is TI if and only if all components of $\mathbf a$ are equal, in which case it reduces to temperature scaling with a class-wise additive bias. MS is TI if and only if $\mathbf W \mathbf 1 = \alpha \mathbf 1$ for some $\alpha \in \mathbb R$, which is equivalent to all rows of $\mathbf W$ summing to the same value. Assuming non-constant logits, LTS is TI if and only if the temperature predictor $\tau_\theta$ is invariant to additive logit shifts (proof in \appref{app:lts_ti}).
However, this property is not guaranteed in \cite{ding2021localtemperaturescaling} since $\tau_\theta$ takes the raw logit field $\mathbf Z$ as input.

Violating translation invariance can lead to undesirable behavior. As an example, consider MS: $g(\mathbf z+c\mathbf 1)=S(\mathbf W \mathbf z+ \mathbf b+ c \mathbf W\mathbf 1)$. 
If $\mathbf W\mathbf 1\notin\mathrm{span}\{\mathbf 1\}$, then $g$ is not TI; in the extreme case $c\to+\infty$, the output distribution converges to a one-hot distribution concentrated on
\(k^\star = \arg\max_k (\mathbf{W}\mathbf{1})_k\), independently of \(S(\mathbf{z})\).
Thus, an arbitrary shift of the logits, which leaves the predictive distribution unchanged, can fully determine the calibrated prediction.

\paragraph{Making calibrators translation-invariant (TI).}
We consider two mechanisms to enforce translation invariance.
The first mechanism constrains the calibrator such that additive logit shifts only induce a constant shift in the pre-softmax outputs. For matrix scaling (MS), this can be achieved by enforcing that all rows of $\mathbf{W}$ have identical sums (this variant is denoted MS$_c$). This constraint can be implemented via an unconstrained parameterization. Specifically, we learn a free matrix $\mathbf{W}_{\mathrm{free}} \in \mathbb{R}^{C \times (C-1)}$ and a scalar $s \in \mathbb{R}$. The full matrix $\mathbf{W} \in \mathbb{R}^{C \times C}$ is obtained by appending a final column to $\mathbf{W}_{\mathrm{free}}$, defined such that each row sums to $s$.


The second mechanism consists of feeding the calibrator with canonical logits instead of their raw representatives. In particular, subtracting the free energy $F(\mathbf z)$ (with a sign convention opposite to some energy-based formulations) from $\mathbf z$ yields a canonical representative of the equivalence class, removing the arbitrary global offset while preserving the predictive distribution:
\begin{equation}
\label{eq:canon_logits}
\log S(\mathbf z) = \mathbf z - F(\mathbf z)\,\mathbf{1},
\end{equation}
\begin{equation}
\label{eq:free_energy}
F(\mathbf z) = \log \sum_{c=1}^C e^{z_c}.
\end{equation}
\Eqref{eq:canon_logits} shows that feeding a calibrator with $\log S(\mathbf z)$ is equivalent to subtracting the free energy $F(\mathbf z)$ from $\mathbf z$, both ensuring translation invariance.

This viewpoint recovers Dirichlet calibration (DC)~\citep{kull2019beyond} as MS applied to
canonical logits:
\begin{equation}
g^{\mathrm{DC}}_{W,b}(\mathbf{z})
=
S\bigl(\mathbf{W}\log S(\mathbf{z})+\mathbf{b}\bigr),
\end{equation}
and therefore DC is TI by construction. The same idea applies
to LTS. The temperature predictor $\tau_\theta$ can be made invariant to additive logit shifts by using log-probabilities as input instead of raw logits:
\begin{equation}
g_{\theta}^{\mathrm{LTS}}(\mathbf z^v)
=
S\!\left(
\mathbf z^v /
\tau_\theta\bigl(v; \log S(\mathbf Z), \xi\bigr)
\right).
\end{equation}




\subsection{Argmax/Order Preservation in Calibration}
\revdel{Since calibration and segmentation objectives are not fully aligned, improvements}\rev{Because post-hoc calibration is fitted by minimizing cross-entropy on a small held-out set, improvements} in calibration may come at the expense of segmentation accuracy.
It remains an open question whether enforcing argmax preservation during calibration limits the achievable calibration gains.
To investigate this question, we construct argmax- and order-preserving counterparts of standard calibrators to quantify the resulting calibration–segmentation trade-off.
Among the standard multiclass post-hoc calibrators presented in Eqs.~\ref{eq:calib-ts}--\ref{eq:calib-lts}, only TS, ETS, and LTS are inherently argmax-preserving and order-preserving. When argmax- or order-preserving constraints are imposed, VS and MS collapse to TS (\appref{app:collapse}). Consequently, comparisons between VS/MS and TS do not isolate the effect of enforcing argmax preservation, as they simultaneously involve a severe reduction in model expressiveness.

To obtain meaningful argmax- and order-preserving counterparts of standard calibrators, one must ensure that the imposed constraints do not collapse the model to overly restrictive forms.
This requires introducing additional flexibility beyond global transformations.
We introduce a conditional calibration framework in which each class is equipped with a dedicated calibrator, selected at inference time according to the argmax of the input logits.
Class-conditioning alone does not guarantee argmax or order preservation, since a selected calibrator may still alter the ranking of the logits. We therefore enforce argmax or order preservation within each class-specific calibrator, which ensures that the resulting global model inherits the same property.

\subsubsection{Enforcing Argmax and Order Preservation for Conditional Class-wise Calibration}
\label{sec:apop}
\paragraph{General case.}
Let $\Psi_1$ and $\Psi_2$ denote the argmax and order cones, respectively:
\begin{align*}
\Psi_1 &= \{ \mathbf{z} \in \mathbb{R}^C \mid z_1 \ge z_k,\ \forall k \neq 1 \}, \\
\Psi_2 &= \{ \mathbf{z} \in \mathbb{R}^C \mid z_1 \ge z_2 \ge \cdots \ge z_C \}.
\end{align*}

For any permutation $\pi$, we denote by $\mathbf v[\pi]$ the vector obtained by reordering $\mathbf v$ according to $\pi$:
$$
\mathbf v[\pi]
=
\bigl(v_{\pi(1)},\ldots,v_{\pi(C)}\bigr).
$$
We introduce a permutation $\pi_{\mathbf z}$ associated with each input $\mathbf z$, which induces the canonical reordering
$
\mathbf z^{\mathrm{can}}=\mathbf z[\pi_{\mathbf z}],
$
where $\pi_{\mathbf z}$ is chosen such that $\mathbf z^{\mathrm{can}}\in\Psi_1$ or $\mathbf z^{\mathrm{can}}\in\Psi_2$, depending on whether argmax or order preservation is enforced.
In the argmax-preserving case ($\Psi_1$), $\pi_{\mathbf z}$ exchanges the first index with the index of $\arg\max_k z_k$, while leaving all other indices unchanged.
In the order-preserving case ($\Psi_2$), $\pi_{\mathbf z}$ is defined such that the components of $\mathbf z^{\mathrm{can}}$ are sorted in non-increasing order.

Let
$
i(\mathbf z)=\pi_{\mathbf z}(1)
$
denote the class occupying the first position in the canonical ordering.
The corresponding class-wise calibrator $g_{i(\mathbf z)}$ is then applied to $\mathbf z^{\mathrm{can}}$, producing calibrated probabilities in canonical order.
The final calibrated probabilities are obtained by applying the inverse permutation:
\[
g(\mathbf z)
=
\left(
g_{i(\mathbf z)}
\left(
\mathbf z[\pi_{\mathbf z}]
\right)
\right)
[\pi_{\mathbf z}^{-1}],
\qquad
i(\mathbf z)=\pi_{\mathbf z}(1).
\]

Let $h_i : \mathbb{R}^C \to \mathbb{R}^C$ denote the pre-softmax transformation associated with the $i$-th class-wise calibrator, so that $g_i = S\circ h_i$. The conditional class-wise calibration is argmax-preserving (resp. order-preserving) if $\Psi_1$ (resp. $\Psi_2$) is invariant under $h_i$, i.e., $h_i(\mathbf{z}) \in \Psi_1 \quad \forall \mathbf{z} \in \Psi_1$ (resp. $\Psi_2$).

The key advantage of the conditional formulation is that preservation constraints are enforced in a fixed canonical space, independent of the original permutation of logits. 
This is less restrictive than enforcing argmax or order preservation directly in the original logit space, where the location of the maximum and the ordering pattern may vary across samples.
Note that this leads to a difference in the information available to the class-wise calibrators. In the argmax-preserving setting, $g_i$ operates on inputs expressed in a fixed canonical order, where the mapping between positions and class identities remains consistent across samples. In contrast, in the order-preserving setting, only the first position is consistently associated with the selected class $i$, whereas the remaining positions are associated with ranks rather than fixed class identities. In principle, one could consider extending the order-preserving formulation by learning separate calibrators for different ordering patterns, thereby explicitly modeling distinct rank configurations. In practice, however, this would lead to a combinatorial explosion in the number of possible ordering patterns, making it intractable.

\paragraph{Argmax preservation for conditional class-wise calibration under MS.}

We derive the constraint for a single class-wise affine map and omit the class index $i$ for readability.
We first enforce set invariance of $\Psi_1$ under MS in the absence of bias: $h(\mathbf{z}) = \mathbf{W}\mathbf{z}$. We introduce the transformed coordinates $\mathbf{x} = \mathbf{G}\mathbf{z}$, where
\begin{equation}
\label{eq:G}
\mathbf{G} =
\begin{pmatrix}
1 & 0 & 0 & \cdots & 0\\
1 & -1 & 0 & \cdots & 0\\
1 & 0 & -1 & \cdots & 0\\
\vdots & & \ddots & \ddots & \vdots\\
1 & 0 & \cdots & 0 & -1
\end{pmatrix}
\in \mathbb{R}^{C \times C}.
\end{equation}

In this coordinate system, the argmax cone $\Psi_1$ is mapped to
\[
\Psi_3 = \{ \mathbf{x} \in \mathbb{R}^C \mid x_i \ge 0, i \ge 2 \}.
\]

Invariance of $\Psi_1$ under $\mathbf{W}$ is equivalent to invariance of $\Psi_3$ under  $\mathbf{W}' = \mathbf{G}\mathbf{W}\mathbf{G}^{-1}$, which holds if and only if
\[
\mathbf{W}' =
\begin{pmatrix}
a & \mathbf{r}^\top \\
\mathbf{0} & \mathbf{A}
\end{pmatrix},
\quad
a \in \mathbb{R},\;
\mathbf{r} \in \mathbb{R}^{C-1},\;
\mathbf{0} \in \mathbb{R}^{C-1},\;
\mathbf{A} \in \mathbb{R}_+^{(C-1)\times(C-1)}.
\]

We further extend the framework to the biased case ($\mathbf{b} \neq \mathbf{0}$). Since the bias term directly shifts the logits, preserving the argmax-cone structure requires the bias vector itself to lie in the same cone, i.e., $\mathbf{b} \in \Psi_1$.

An important property of the resulting calibration method is its translation invariance, which can be verified by checking that $\mathbf{W} \mathbf{1}_C = a \mathbf{1}_C$.
Moreover, we derive from the previous results a fully differentiable and unconstrained parameterization.
We parameterize a free matrix $\tilde{\mathbf{A}} \in \mathbb{R}^{(C-1)\times(C-1)}$ and enforce non-negativity via a softplus transformation:
$
\mathbf{A}_{ij} = \mathrm{softplus}(\tilde{\mathbf{A}}_{ij})$. We then construct \(\mathbf{W}' \in \mathbb{R}^{C\times C}\) by embedding \(\mathbf{A}\) into the lower-right block, setting the first column (except for \(\mathbf W'_{11}\)) to zero, and parameterizing the first row as an unconstrained vector in \(\mathbb{R}^C\). Then, $\mathbf W$ is set to 
$\mathbf G^{-1} \mathbf W'\mathbf G$.
For the bias, we parameterize $\tilde{\mathbf{b}} \in \mathbb{R}^C$ and enforce ordering via $
\mathbf{b}_1 = \tilde{\mathbf{b}}_1$ and 
$\mathbf{b}_i = \mathbf{b}_1 - \mathrm{softplus}(\tilde{\mathbf{b}}_i)$ if $i \ge 2
$.

\paragraph{Order preservation for conditional class-wise calibration under MS.}
Enforcing invariance of $\Psi_2$ under MS follows exactly  the same construction as for $\Psi_1$ with:
\[
\mathbf{G} =
\begin{pmatrix}
1 & 0 & 0 & \cdots & 0\\
1 & -1 & 0 & \cdots & 0\\
0 & 1 & -1 & \cdots & 0\\
\vdots & & \ddots & \ddots & \vdots\\
0 & 0 & \cdots & 1 & -1
\end{pmatrix}
\in \mathbb{R}^{C \times C}.
\]

The bias vector must now satisfy $\mathbf{b} \in \Psi_2$. To enforce this constraint, we parameterize a free vector $\tilde{\mathbf{b}} \in \mathbb{R}^C$ and define
$
\mathbf{b}_1 = \tilde{\mathbf{b}}_1$, 
$\mathbf{b}_i = \mathbf{b}_{i-1} - \mathrm{softplus}(\tilde{\mathbf{b}}_i)$ if $i \ge 2,
$
which guarantees
$
\mathbf{b}_1 \ge \mathbf{b}_2 \ge \cdots \ge \mathbf{b}_C
$.

\subsection{Instantiation of Segmentation/Calibration Pipelines}
\label{sec:pipelines}

Logits provided to the calibrator are produced by a segmentation model. In this work, we consider deep ensembles~\citep{lakshminarayanan_simple_2017}, which are widely adopted in semantic segmentation to improve performance and predictive robustness. They typically yield higher Dice scores (DSC) and improved reliability compared to single networks~\citep{lakshminarayanan_simple_2017, fort_deep_2020, ovadia_can_2019}. However, improved segmentation performance does not necessarily translate into well-calibrated confidence estimates, and ensemble predictions are often still miscalibrated~\citep{rahaman_uncertainty_2021}.

An ensemble consists of $M$ independently trained segmentation models. At voxel $v$, model $m$ produces logits $\mathbf{z}_m(v) \in \mathbb{R}^C$. Ensemble predictions are aggregated by a pooling operator acting either on logits $\{\mathbf{z}_m(v)\}_{m=1}^M$ or on probabilities $\{\mathbf{p}_m(v)\}_{m=1}^M$, yielding pooled logits $\mathbf{z}(v)$ or pooled probabilities $\mathbf{p}(v)$ that are passed to the calibrator. When pooling is performed in probability space, we provide $\log \mathbf{p}(v)$ to the calibrator, ensuring consistency with the logit-based formulation.

We consider two standard pooling strategies: probability averaging, also known as mixture-of-experts fusion, and logit averaging, which induces a product-of-experts--type aggregation and penalizes inter-model disagreement more strongly. Pipelines based on probability averaging are denoted $g(\bar{\mathbf{p}})$, whereas those based on logit averaging are denoted $g(\bar{\mathbf{z}})$. Both pooling operators are invariant to member-wise additive logit shifts: adding $\delta_m\mathbf{1}$ to all logits of model $m$ leaves the resulting pooled probabilities unchanged. Moreover, $g(\bar{\mathbf{p}})$ pipelines are TI by construction, since the calibrator operates on log-probabilities rather than logits, which  guarantees translation invariance (Sec.~\ref{sec:translation}).

We next introduce the pipelines considered in our experiments.

\paragraph{Uncalibrated baselines.}
To assess the intrinsic impact of calibration, we consider three uncalibrated baselines: a single segmentation model, denoted $p_0$, and a deep ensemble using either logit averaging ($\bar{\mathbf{z}}$) or probability averaging ($\bar{\mathbf{p}}$) for pooling.

\paragraph{Translation invariance.}
To investigate the effect of translation invariance, we compare the non-TI pipeline MS($\bar{\mathbf{z}}$) with two TI counterparts: MS$_c$($\bar{\mathbf{z}}$) and DC($\bar{\mathbf{z}}$). \revdel{MS$_c$ enforces translation invariance in logit space via a row-sum constraint on the weight matrix, while DC($\bar{\mathbf{z}}$) can be interpreted as applying MS to canonical log-probabilities $\log S(\bar{\mathbf{z}})$ (cf. Section~3.1).} \rev{MS$_c$ and DC implement translation invariance in two different ways (cf. Section~\ref{sec:translation}): MS$_c$ enforces it directly through a constraint on the affine transformation while retaining the original logit representation, whereas DC obtains it by mapping the logits to their canonical log-softmax representation (DC can be viewed as applying MS to canonical log probabilities). Thus, MS vs.\ MS$_c$ provides a controlled comparison that isolates the effect of translation invariance without changing the input representation, while MS vs.\ DC provides a controlled comparison of the effect of the representation space (original vs.\ canonical).}

\rev{MS vs.\ MS$_c$ may therefore be considered the cleanest TI ablation. However, the log-softmax reparameterization is itself a way of making a calibrator translation invariant, so MS vs.\ DC is also a relevant comparison for studying an alternative implementation of translation invariance. For simplicity, in the remainder of the paper, we refer to both MS$_c$ and DC as TI counterparts, while keeping this distinction explicit here.} Note that MS becomes equivalent to DC when pooling is performed in probability space ($\bar{\mathbf{p}}$); therefore, it is not included in this comparison.

For LTS, translation invariance depends on the representation provided to the temperature network $\tau_\theta$, as discussed in Section~\ref{sec:translation}. We consider two variants: the original logit-based formulation of~\cite{ding2021localtemperaturescaling}, which is not TI, and a probability-based variant that enforces it. The first is denoted LTS($\bar{\mathbf{z}}$), and the second LTS($\log S(\bar{\mathbf{z}})$), highlighting that the temperature network operates on canonical log-probabilities of the pooled predictions.

\paragraph{Argmax/Order preservation.}
We denote CDC($\bar{\mathbf{p}}$) and CDC($\bar{\mathbf{z}}$) the conditional class-wise calibration schemes introduced in Section~\ref{sec:apop}, where each class is equipped with a DC calibrator. Consequently, both pipelines are translation-invariant (TI) but neither order- nor argmax-preserving. We compare CDC($\bar{\mathbf{p}}$) and CDC($\bar{\mathbf{z}}$) with their respective order- and argmax-preserving counterparts, constructed using the algorithms presented at the end of Section~\ref{sec:apop}.
For probability pooling, CDC($\bar{\mathbf{p}}$) is compared with CMS$_{\mathrm{op}}$($\bar{\mathbf{p}}$) and CMS$_{\mathrm{ap}}$($\bar{\mathbf{p}}$), where CMS$_{\mathrm{op}}$ enforces order preservation and CMS$_{\mathrm{ap}}$ enforces argmax preservation. Similarly, for logit pooling, CDC($\bar{\mathbf{z}}$) is compared with CMS$_{\mathrm{op}}$($\bar{\mathbf{z}}$) and CMS$_{\mathrm{ap}}$($\bar{\mathbf{z}}$).

Note that CMS$_{\mathrm{op}}$ and CMS$_{\mathrm{ap}}$ are TI, whereas CMS is not. Therefore, comparing CMS$_{\mathrm{op}}$ and CMS$_{\mathrm{ap}}$ with CMS would conflate translation invariance with order or argmax constraints. In contrast, CDC is TI by construction and is thus used as the non–order-/non–argmax-preserving reference.

\paragraph{Standard pipelines.}
We also evaluate standard temperature-based calibrators (TS, ETS), which preserve class ordering and are TI. We further include Dirichlet calibration (DC), which is TI by construction. This allows us to assess both the impact of the pooling operator and the role of calibrator expressiveness on performance.

\section{Experiments and Results}
\label{sec:experiments}

\subsection{Experimental Setup}

\begin{figure*}[tb]
    \centering
    \includegraphics[width=\linewidth]{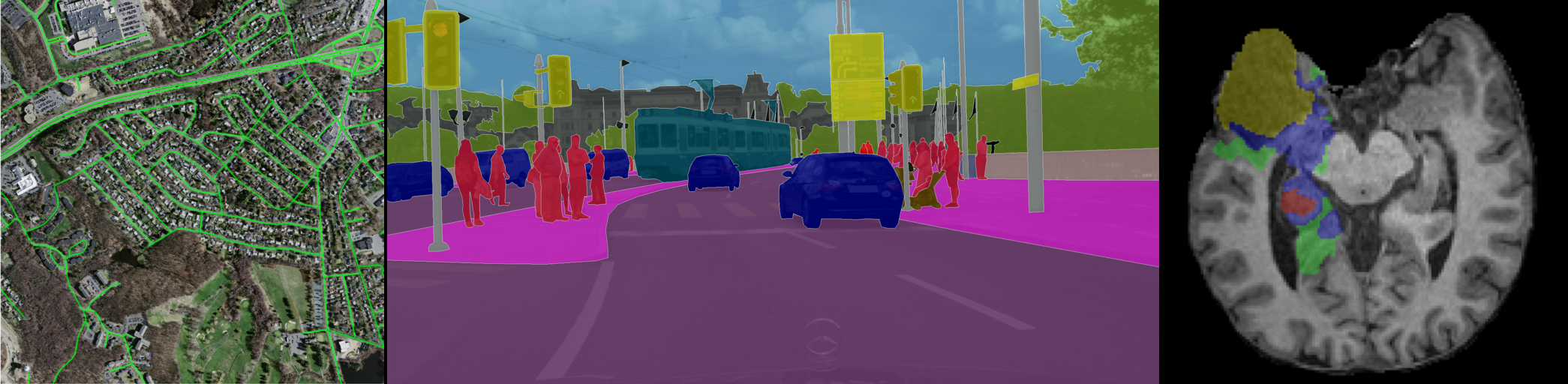}
    \caption{Sample inputs with ground-truth segmentation overlays from Massachusetts Roads, Cityscapes, and BraTS2024 Glioma (left to right). For BraTS, we show a 2D slice from the 3D MRI volume.}
    \label{fig:dataset_examples}
\end{figure*}

\paragraph{Datasets.}
We evaluate the calibration pipelines from Section~\ref{sec:pipelines} on three segmentation benchmarks spanning medical and natural-image domains:
BraTS GLI 2024~\citep{brats2024} (3D brain MRI glioma segmentation),
Cityscapes~\citep{cordts2016cityscapes} (2D urban scene parsing), and
Massachusetts Roads (RoadSeg)~\citep{mnih2013thesis} (2D aerial road extraction).
These datasets vary in modality (MRI vs.\ RGB), dimensionality (3D vs.\ 2D),
and number of classes, thereby enabling evaluation across heterogeneous imaging and semantic settings.
Dataset statistics, split sizes, and the number of classes are summarized in
Table~\ref{tab:datasets}. 
As shown in the table, we use a calibration set of size
$n=50$ for all experiments; a sensitivity analysis (\appref{app:calib_size}) indicates that
calibration metrics largely stabilize for $n \ge 25$.
Representative input images and ground-truth overlays are shown in Figure~\ref{fig:dataset_examples}.

\begin{table*}[tb]
\centering
\footnotesize
\setlength{\tabcolsep}{7pt}
\begin{tabular}{@{}llcccccc@{}}
\toprule
\textbf{Dataset} & \textbf{Modality} & $\bm{C}$ & \textbf{Seg.\ train} &
$\bm{|\mathcal{D}_{\mathrm{cal}}|}$ & $\bm{|\mathcal{D}_{\mathrm{val}}|}$ & $\bm{|\mathcal{D}_{\mathrm{test}}|}$ & \textbf{Pool} \\
\midrule
BraTS GLI 2024~\citep{brats2024}              & 3D MRI & 5  & 1350           & 50 & 27 & 194 & 271 \\
Cityscapes~\citep{cordts2016cityscapes}       & 2D RGB & 20 & 2975           & 50 & 50 & 400 & 500 \\
Massachusetts\ Roads (RoadSeg)~\citep{mnih2013thesis} & 2D RGB & 2  & 566 & 50 & 24 & 169 & 243 \\
\bottomrule
\end{tabular}
\caption{\textbf{Dataset splits and class counts.} Segmentation models are trained
on the official training split for BraTS (\emph{TrainData}, 1350 cases) and
Cityscapes (\emph{train}, 2975 images).
The remaining data form a held-out pool partitioned into
$\mathcal{D}_{\mathrm{cal}}/\mathcal{D}_{\mathrm{val}}/\mathcal{D}_{\mathrm{test}}$:
for BraTS, we use the official additional training release (271 labeled cases); for
Cityscapes we use the official \emph{val} set (test-set labels are withheld); for Massachusetts Roads, we exclude incomplete/occluded images, yielding 809 image--mask pairs, of which 243 form the held-out pool.}
\label{tab:datasets}
\end{table*}


\paragraph{Segmenter training.}
For each dataset, we train a deep ensemble of $M=5$ nnU-Net~\citep{isensee_nnu-net_2021} v2.4.1 models initialized with different random seeds.
The 2D or 3D configuration is selected according to the dataset dimensionality, both using the \texttt{ResEncM} preset and default preprocessing.
Optimization follows the nnU-Net defaults (SGD with momentum, \texttt{polyLR} schedule, deep supervision, and automatic patch and batch-size configuration). \rev{The models use the nnU-Net default compound objective, an equal-weight sum of soft Dice and cross-entropy, rather than Dice alone.}
All ensemble members share identical preprocessing, data augmentation, and optimization settings.
The ensemble is trained once per dataset on the official training split and remains fixed for all subsequent calibration experiments.

\paragraph{Post-hoc calibration protocol.}
Calibration is performed post hoc on the fixed ensemble segmenter.
For each dataset, we perform $R=3$ independent repeats, each using a new random partition of the held-out images (Table~\ref{tab:datasets}) into
calibration, validation, and test subsets.
In each repeat $r$, calibrator parameters are fitted on
$\mathcal{D}^{(r)}_{\mathrm{cal}}$ by minimizing the voxel-wise negative log-likelihood (NLL) of the pooled ensemble predictions with an additional regularization term. When applicable, early stopping and hyperparameter selection use $\mathcal{D}^{(r)}_{\mathrm{val}}$, and final results are reported on $\mathcal{D}^{(r)}_{\mathrm{test}}$.
Details on optimization, regularization, and hyperparameter search are provided in \appref{app:hpo}.

\paragraph{Model parameterization and expressiveness}
A common way to compare the expressiveness of calibration methods is to count trainable parameters. However, this can be partially misleading since most used models are overparameterized due to the softmax’s invariance to additive constants. It is therefore more meaningful to consider the number of identifiable parameters.
For example, in the  MS case, subtracting from each column of $\mathbf{W}$ its first entry and subtracting $b_1$ from $\mathbf{b}$ leaves the induced softmax distribution unchanged. Consequently, without loss of generality, one may fix the first row of $\mathbf{W}$ to zero and set $\mathbf b_1 = 0$.

\begin{table*}[t]
\centering
\footnotesize
\setlength{\tabcolsep}{8pt}
\small
\begin{tabular}{@{}l cc cc ccc@{}}
\toprule
\textbf{Calibrator} &
\textbf{TI} &
\makecell{\textbf{Decision}\\\textbf{preservation}} &
\multicolumn{1}{c}{\textbf{Optimized}} &
\multicolumn{1}{c}{\textbf{Identifiable}} &
\multicolumn{1}{c}{\makecell{\textbf{RoadSeg}\\$(C{=}2)$}} &
\multicolumn{1}{c}{\makecell{\textbf{BraTS}\\$(C{=}5)$}} &
\multicolumn{1}{c}{\makecell{\textbf{Cityscapes}\\$(C{=}20)$}}
\\ \midrule
\multicolumn{8}{@{}l}{\textit{Global calibrators}} \\
TS                & \cmark & Order   & $1$        & $1$        & $1/1$     & $1/1$     & $1/1$ \\
ETS               & \cmark & Order   & $4$        & $3$        & $4/3$     & $4/3$     & $4/3$ \\
MS                & \xmark & \xmark  & $C(C+1)$   & $C^2-1$    & $6/3$     & $30/24$   & $420/399$ \\
MS$_{\mathrm{c}}$ & \cmark & \xmark  & $C^2+1$    & $C(C-1)$   & $5/2$     & $26/20$   & $401/380$ \\
DC                & \cmark & \xmark  & $C(C+1)$   & $C^2-1$    & $6/3$     & $30/24$   & $420/399$ \\
\multicolumn{8}{@{}l}{\textit{Spatially adaptive or class-conditional calibrators}} \\
LTS               & \xmark\textsuperscript{\dag} & Order & $7(25C+1)$ & $7(25C+1)$ & $357/357$ & $882/882$ & $3\,507/3\,507$ \\
CDC               & \cmark & \xmark     & $C^2(C+1)$ & $C(C^2-1)$ & $12/6$    & $150/120$ & $8\,400/7\,980$ \\
CMS$_{\mathrm{ap}}$ & \cmark & Argmax & $C(C^2+1)$ & $C^2(C-1)$ & $10/4$  & $130/100$ & $8\,020/7\,600$ \\
CMS$_{\mathrm{op}}$ & \cmark & Order  & $C(C^2+1)$ & $C^2(C-1)$ & $10/4$  & $130/100$ & $8\,020/7\,600$ \\
\bottomrule
\end{tabular}
\caption{\textbf{Calibrator properties, optimized parameters, and identifiable degrees of freedom.}
``TI'' indicates a translation-invariant calibrator; ``Decision~preservation'' indicates whether the calibrator preserves the predicted decision (Argmax), the full class ordering (Order), or neither (\xmark).
``Optimized'' counts fitted scalars; ``Identifiable'' removes continuous softmax-gauge
directions.
The last three columns report the optimized and identifiable counts for each calibrator and dataset.
\textsuperscript{\dag}LTS is TI only for our variant LTS$(\log S(\bar z))$, in which the temperature network operates on canonical log-probabilities, not for the original formulation LTS$(\bar z)$.}
\label{tab:num_params}
\end{table*}

Table~\ref{tab:num_params} reports both the total number of parameters and the number of identifiable parameters for each calibrator. In particular, the identifiable parameter counts  exploit the fact that any MS-based calibrator admits an equivalent parameterization with $\mathbf b_1 = 0$ and a zero first row in $\mathbf{W}$. For CMS$_{ap}$ and CMS$_{op}$, the first row of the matrix $\mathbf{A}$ from which $\mathbf{W}$ is computed can also be fixed to zero without loss of generality.

In practice, we retain the overparameterized parameterizations during optimization and account for these redundancies only when comparing model expressiveness. Importantly, overparameterized models are necessary during optimization due to the regularization term used during training (cf. Appendix~\ref{app:hpo}). We use a Dirichlet-style regularizer, described in~\cite{kull2019beyond}, that penalizes off-diagonal weights and large biases via an $\ell_2$ norm. Under such a regularization, the identity mapping is not penalized; however, in the reduced parameterization presented above, the identity is not representable, and its equivalent representation has a first column given by $(0, -1, -1, \dots)^\top$, which is strongly penalized. This leads to different regularization behavior depending on the chosen parameterization.

\rev{The number of optimized (or identifiable) parameters in CDC, CMS$_{ap}$, and CMS$_{op}$ grows as $\mathcal{O}(C^3)$, making these calibrators suitable primarily for low to moderate numbers of classes.}

\paragraph{Evaluation.}
Segmentation accuracy is measured with the multiclass Dice score (DSC), which allows us to quantify whether calibration affects segmentation performance. For methods that may change the predicted class, we further report the class-flip rate, defined as the percentage of voxels whose predicted class changes after calibration. Calibration is assessed with negative log-likelihood (NLL), expected calibration error (ECE)~\citep{guo_calibration_2017}, boundary-aware ECE (BA-ECE)~\citep{zeevi_spatially-aware_2025}, and Average Calibration Error (ACE)~\citep{kahl2024values}. NLL is a strictly proper scoring rule and corresponds to the loss used to fit the calibration models. ECE is computed with $B=50$ fixed-width confidence bins; a sensitivity analysis over $B\in\{20,30,40,50\}$ shows that ECE values change only marginally and that the relative ranking across methods is stable (\appref{app:bin}). BA-ECE evaluates calibration as a function of distance to the ground-truth boundary; details on distance bands and weighting are provided in \appref{app:spatial}. Since ECE weights bin-wise calibration gaps by bin occupancy, it can be dominated by highly populated confidence ranges. We therefore also report ACE, which averages calibration gaps uniformly over non-empty bins; we use $B=15$, with a binning sensitivity analysis also reported in~\appref{app:bin}.
All metrics, except ACE, are computed for each image in the held-out test set and then averaged across repeats. We report 95\% confidence intervals obtained via a two-level hierarchical bootstrap~\citep{efron1994introduction} with $N_{\mathrm{boot}}=2000$ replicates, resampling repeats first and then cases within each selected repeat. \rev{For matched comparisons, we also compute paired 95\% confidence intervals for the difference between the metrics obtained with the two methods. We resample repeats, then resample cases jointly for both methods, with $N_{\mathrm{boot}}=10{,}000$ replicates. We consider a difference to be significant when its 95\% confidence interval excludes zero.} 
For ACE, image-level estimates can be unstable because some confidence bins may contain very few voxels.
We therefore compute ACE by pooling voxels within each repeat and report the resulting repeat-level values as mean [min, max].


\subsection{Pooled Logit Offset Variability}
\label{sec:logit_offset_real}
As discussed in Sec.~\ref{sec:translation}, segmenter logits can be shifted by an arbitrary constant without affecting the resulting softmax probabilities. To analyze whether such shifts occur in practice,
we compute the free energy (cf.~\Eqref{eq:free_energy}).
Since such offsets can influence non-TI calibrators, this provides a practical diagnostic for whether enforcing translation invariance can matter. It is not, by itself, a predictor of the empirical gain from enforcing invariance: a non-TI calibrator may still learn an approximately invariant solution if the calibration data expose the relevant offsets. For each held-out image, we compute the spatial standard deviation of the pooled-logit free energy, $\mathrm{Std}_v(F_{\mathrm{pool}}(v))$. Table~\ref{tab:offset_stats} reports the mean across images for ensembles with $M{=}5$ models. Spatial variability is substantial for all datasets and is markedly larger for BraTS and Cityscapes than for RoadSeg. The same trend holds for smaller ensembles (see \appref{app:ensemble_size}).

\begin{table}[tb]
\centering
\small
\setlength{\tabcolsep}{12pt}
\renewcommand{\arraystretch}{0.85}
\begin{tabular}{l c r}
\toprule
\textbf{Dataset} & \textbf{$n_{\text{cases}}$} &
\textbf{$\mathbb{E}[\mathrm{Std}_v(F_{\mathrm{pool}}(v))]$} \\
\midrule
RoadSeg                      & 243 & 1.08 {\scriptsize[0.73, 1.06, 1.34]} \\
Cityscapes                   & 500 & 2.54 {\scriptsize[2.30, 2.46, 2.71]} \\
BraTS                        & 271 & 8.77 {\scriptsize[8.56, 8.69, 8.87]} \\
\bottomrule
\end{tabular}
\caption{%
Spatial variability of the pooled-logit free energy 
$F_{\mathrm{pool}}(v) = \log \sum_{c=1}^{C} \exp(z_{\mathrm{pool},c}(v))$,
where $z_{\mathrm{pool}}(v)$ denotes the pooled logits at voxel $v$.
Statistics are computed from nnU-Net ensembles with $M=5$.
For each dataset, we report the mean of $\mathrm{Std}_v(F_{\mathrm{pool}}(v))$ across held-out cases, with interquartile range
$[p_{25}, p_{50}, p_{75}]$.
}
\label{tab:offset_stats}
\end{table}

\begin{figure}[tb]
\centering
\includegraphics[width=\textwidth]{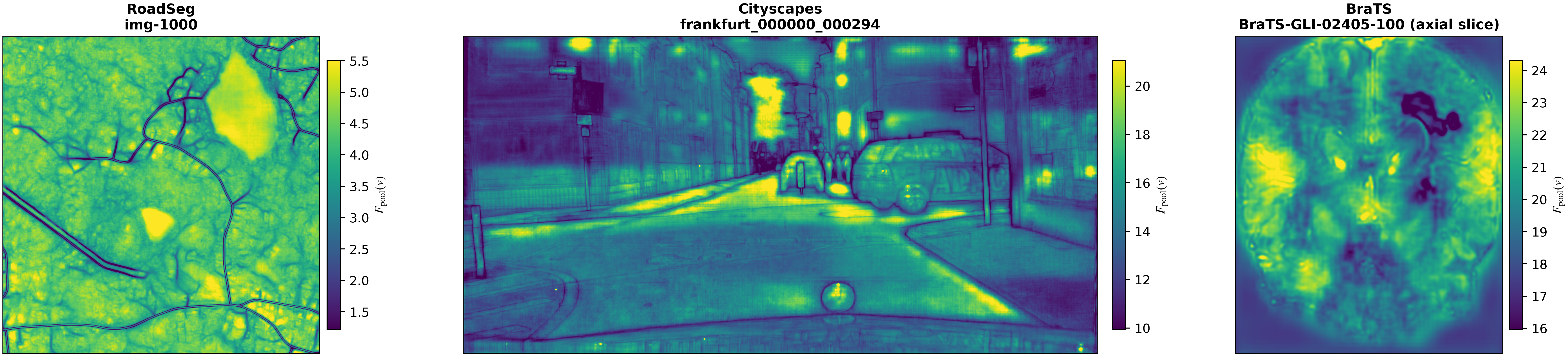}
\caption{Spatial maps of the pooled-logit free energy 
$F_{\mathrm{pool}}(v) = \log \sum_{c=1}^{C} \exp(z_{\mathrm{pool},c}(v))$
for representative test images from RoadSeg, Cityscapes, and BraTS.
}
\label{fig:offset_maps}
\end{figure}

Figure~\ref{fig:offset_maps} shows representative spatial maps of $F_{\mathrm{pool}}(v)$. 
These maps illustrate the spatial variability of the pooled-logit free energy across datasets.

\subsection{Comparison of Post-Hoc Calibration Pipelines}
\label{sec:overall_results}

We evaluate the calibration pipelines (Sec.~\ref{sec:pipelines}) across the three segmentation benchmarks. Table~\ref{tab:results} summarizes the results in terms of calibration and segmentation metrics. For conciseness, only point estimates are reported; corresponding intervals are provided in \appref{app:results}.
Note that the trends observed in-distribution persist under corruption-based covariate shift on Cityscapes, and are often amplified under stronger perturbations (\appref{app:ood}):

\begin{table*}[t]
\centering
\small
\setlength{\tabcolsep}{2pt}
\renewcommand{\arraystretch}{1.3}
\resizebox{\textwidth}{!}{%
\begin{tabular}{@{}l
c c c c c c
c c c c c c
c c c c c c
@{}}
\toprule
& \multicolumn{6}{c}{\textbf{RoadSeg}~\cite{mnih2013thesis}} & \multicolumn{6}{c}{\textbf{BraTS}~\cite{brats2024}} & \multicolumn{6}{c}{\textbf{Cityscapes}~\cite{cordts2016cityscapes}} \\
\cmidrule(lr){2-7}\cmidrule(lr){8-13}\cmidrule(lr){14-19}
\textbf{Pipeline} &
{\textbf{NLL}$\downarrow$} & {\textbf{DSC}$\uparrow$} & {\textbf{ECE}$\downarrow$} & {\textbf{BA-ECE}$\downarrow$} & {\textbf{ACE}$\downarrow$} & {\textbf{Flip}$\downarrow$} &
{\textbf{NLL}$\downarrow$} & {\textbf{DSC}$\uparrow$} & {\textbf{ECE}$\downarrow$} & {\textbf{BA-ECE}$\downarrow$} & {\textbf{ACE}$\downarrow$} & {\textbf{Flip}$\downarrow$} &
{\textbf{NLL}$\downarrow$} & {\textbf{DSC}$\uparrow$} & {\textbf{ECE}$\downarrow$} & {\textbf{BA-ECE}$\downarrow$} & {\textbf{ACE}$\downarrow$} & {\textbf{Flip}$\downarrow$} \\
\midrule

\multicolumn{19}{@{}l}{\textbf{(a) Uncalibrated baselines}}\\
\addlinespace[1pt]
$p_0$ & 0.091& 76.3& 0.017& 0.094& 0.119& \NA& 0.524& 75.2& 0.340& 0.049& 0.112& \NA& 0.485& 71.7& 0.063& 0.182& 0.125& \NA \\
$\bar{p}$ & 0.085& \bfseries 76.7& 0.015& 0.087& 0.105& \NA& 0.522& 76.0& 0.340& 0.041& 0.087& \NA& 0.377& 78.1& 0.044& 0.136& 0.079& \NA \\
$\bar{z}$ & 0.088& \bfseries 76.7& 0.016& 0.091& 0.117& \NA& 0.523& \bfseries 76.9& 0.340& 0.046& 0.096& \NA& 0.434& \bfseries 79.4& 0.058& 0.170& 0.122& \NA \\
\addlinespace[1pt]
\midrule

\multicolumn{19}{@{}l}{\textbf{(b) Translation invariance: MS vs.\ MS$_c$ and DC}}\\
\addlinespace[1pt]
\rowcolor{tirow} MS($\bar{z}$)\ninv & 0.068& 76.5& 0.010& 0.056& 0.053& 0.2& 0.008& 68.4& 0.001& 0.043& 0.069& \bfseries 0.1& 0.404& 73.4& 0.045& 0.156& 0.072& 3.8 \\
\rowcolor{tirow} DC($\bar{z}$) & 0.067\sig& \bfseries 76.7\sig& 0.008\sig& 0.047\sig& 0.040& \bfseries 0.1& 0.008& 67.6& 0.001& 0.040& 0.065& \bfseries 0.1& 0.324\sig& 70.4& 0.035\sig& 0.096\sig& 0.040& \bfseries 2.3  \\
\rowcolor{tirow} MS$_c$($\bar{z}$) & 0.068& 76.5\sig& 0.010& 0.057& 0.055& 0.2& \bfseries 0.007& 70.2& 0.001& 0.033& 0.041& \bfseries 0.1& 0.328\sig& 72.8& 0.034\sig& 0.098\sig& 0.046& 2.4  \\
\addlinespace[1pt]
\midrule

\multicolumn{19}{@{}l}{\textbf{(c) Translation invariance: LTS with logits vs.\ log-probabilities}}\\
\addlinespace[1pt]
\rowcolor{tirow} LTS($\bar{z}$)\ninv & 0.067& \bfseries 76.7& 0.004& 0.047& 0.024& \NA& 0.018& \bfseries 76.9& 0.012& 0.134& 0.237& \NA& 0.344& \bfseries 79.4& 0.016& 0.075& 0.039& \NA \\
\rowcolor{tirow} LTS($\log S(\bar{z})$) & 0.066& \bfseries 76.7& \bfseries 0.002\sig& \bfseries 0.040& \bfseries 0.007& \NA& 0.014& \bfseries 76.9& 0.008& 0.090& 0.216& \NA& 0.342& \bfseries 79.4& \bfseries 0.010& 0.075& 0.029& \NA  \\
\addlinespace[1pt]
\midrule

\multicolumn{19}{@{}l}{\textbf{(d) Decision preservation: CDC vs CMS}}\\
\addlinespace[1pt]
\rowcolor{dprow} CDC($\bar{p}$) & \bfseries 0.065\sig& 75.7& 0.006\sig& 0.043& 0.013& 0.5& 0.008& 72.0& 0.001& 0.034& 0.046& \bfseries 0.1& 0.335& 69.3& 0.032& 0.108& 0.052& 5.1  \\
\rowcolor{dprow} CMS$_{ap}$($\bar{p}$) & 0.067& \bfseries 76.7\sig& 0.007& 0.046& 0.036& \NA& \bfseries 0.007\sig& 76.0\sig& 0.001& 0.029\sig& 0.044& \NA& \bfseries 0.314\sig& 78.1\sig& 0.035& 0.086\sig& 0.041& \NA  \\
\rowcolor{dprow} CMS$_{op}$($\bar{p}$) & 0.067& \bfseries 76.7\sig& 0.007& 0.046& 0.037& \NA& \bfseries 0.007\sig& 76.0\sig& 0.001& \bfseries 0.028\sig& 0.065& \NA& 0.326\sig& 78.1\sig& 0.035& 0.071\sig& 0.032& \NA  \\
\cmidrule(lr){2-19}
\rowcolor{dprow} CDC($\bar{z}$) & 0.066\sig& 75.7& 0.006\sig& 0.041\sig& 0.012& 0.6& 0.008& 74.0& 0.001& 0.035& \bfseries 0.033& \bfseries 0.1& 0.341& 74.2& 0.035& 0.115& 0.067& 5.0  \\
\rowcolor{dprow} CMS$_{ap}$($\bar{z}$) & 0.067& \bfseries 76.7\sig& 0.008& 0.047& 0.043& \NA& \bfseries 0.007& \bfseries 76.9\sig& 0.001& 0.031\sig& 0.041& \NA& 0.327\sig& \bfseries 79.4\sig& 0.038& 0.091\sig& 0.081& \NA  \\
\rowcolor{dprow} CMS$_{op}$($\bar{z}$) & 0.067& \bfseries 76.7\sig& 0.007& 0.047& 0.043& \NA& \bfseries 0.007& \bfseries 76.9\sig& 0.001& 0.030\sig& 0.055& \NA& 0.339& \bfseries 79.4\sig& 0.035& 0.079\sig& 0.041& \NA  \\
\addlinespace[1pt]
\midrule

\multicolumn{19}{@{}l}{\textbf{(e) Standard temperature-based calibrators (order-preserving, TI): TS/ETS}}\\
\addlinespace[1pt]
\rowcolor{poolrow} TS($\bar{p}$) & 0.068& \bfseries 76.7& 0.008& 0.059& 0.056& \NA& 0.025& 76.0& 0.005& 0.046& 0.146& \NA& 0.326& 78.1& 0.011& \bfseries 0.069& \bfseries 0.025& \NA \\
\rowcolor{poolrow} TS($\bar{z}$) & 0.068& \bfseries 76.7& 0.009& 0.061& 0.065& \NA& 0.029& \bfseries 76.9& 0.007& 0.048& 0.167& \NA& 0.338& \bfseries 79.4& 0.012& 0.079& 0.039& \NA \\
\cmidrule(lr){2-19}
\rowcolor{poolrow} ETS($\bar{p}$) & 0.068& \bfseries 76.7& 0.008& 0.059& 0.055& \NA& 0.010& 76.0& \bfseries 0.000& 0.047& 0.166& \NA& 0.326& 78.1& \bfseries 0.010& 0.071& 0.026& \NA \\
\rowcolor{poolrow} ETS($\bar{z}$) & 0.068& \bfseries 76.7& 0.009& 0.061& 0.065& \NA& 0.010& \bfseries 76.9& \bfseries 0.000& 0.048& 0.187& \NA& 0.338& \bfseries 79.4& 0.013& 0.079& 0.039& \NA \\
\addlinespace[1pt]
\midrule

\multicolumn{19}{@{}l}{\textbf{(f) Standard affine calibrator (non order-preserving, TI): DC}}\\
\addlinespace[1pt]
\rowcolor{poolrow} DC($\bar{p}$) & 0.067& 76.6& 0.008& 0.046& 0.033& \bfseries 0.1& 0.008& 68.4& 0.001& 0.037& 0.057& \bfseries 0.1& 0.319& 64.8& 0.033& 0.098& 0.037& 3.2 \\
\rowcolor{poolrow} DC($\bar{z}$) & 0.067& \bfseries 76.7& 0.008& 0.047& 0.040& \bfseries 0.1& 0.008& 67.6& 0.001& 0.040& 0.065& \bfseries 0.1& 0.324& 70.4& 0.035& 0.096& 0.040& \bfseries 2.3 \\
\addlinespace[1pt]

\bottomrule
\end{tabular}%
}
\caption{\textbf{Calibration results.} DSC and Flip are reported in \%.
\textemdash\ indicates not applicable; $\times$ denotes non--translation-invariant pipelines.
\textbf{Bold} indicates the best value per dataset and metric\rev{\ and carries no claim of significance}. \rev{Each block pairs a base method with its counterparts. An $\ast$ on the base method indicates that it outperforms all of its counterparts; an $\ast$ on a counterpart indicates that it outperforms the base method. Here, ``outperforms'' means that the 95\% paired confidence interval for the metric difference excludes zero in favor of the indicated method (\appref{app:results}). ACE and Flip are not paired endpoints and therefore carry no stars.}
Light row colors identify the matched comparison family: blue for translation invariance, green for decision preservation, and beige for standard calibrators.
}
\label{tab:results}
\end{table*}

\paragraph{\textbf{1. Ensembles improve segmentation but do not reliably fix calibration. (Table~\ref{tab:results}(a))}}
Ensemble pooling consistently improves segmentation accuracy over a single model, although the magnitude of the gain depends on the dataset.
On Cityscapes, DSC increases from $71.7$ ($p_0$) to $78.1$ with probability pooling ($\bar p$) and $79.4$ with logit pooling ($\bar z$).
On BraTS, the improvement is more moderate ($75.2$ to $76.0$--$76.9$), while on RoadSeg it is marginal ($76.3$ to $76.7$).
Logit pooling yields slightly higher DSC than probability pooling on BraTS and Cityscapes, with negligible differences on RoadSeg.

For calibration, ensemble pooling improves or leaves metrics essentially unchanged, but the magnitude of the gains depends on the dataset and pooling strategy. Improvements are marginal on RoadSeg, more pronounced on Cityscapes with probability pooling, and mixed across metrics on BraTS. 

Overall, compared with a single model, ensembling improves or preserves segmentation and does not degrade calibration metrics in our experiments. However, the calibration gains remain modest compared with those obtained by dedicated post-hoc calibration methods, motivating the calibration pipelines evaluated next.

\paragraph{\textbf{2. TI pipelines provide representation-consistent calibration.}}

We next evaluate whether enforcing translation invariance (TI) improves calibration in matched pipeline comparisons.
Calibration quality also depends on other factors, in particular calibrator expressivity.
It is therefore not surprising that, outside matched pairs, a non-TI method may outperform a TI method. 

We first compare matrix scaling MS($\bar z$) with its two TI counterparts MS$_c$($\bar z$) and DC($\bar z$) (Table~\ref{tab:results}(b)).
Cityscapes gives the clearest effect: relative to MS($\bar z$), both TI variants reduce NLL (from $0.404$ to $0.324$--$0.328$), BA-ECE (from $0.156$ to $0.096$--$0.098$), and ACE (from $0.072$ to $0.040$--$0.046$).
On BraTS, where ECE is dominated by the background class, class- and boundary-aware metrics show the same trend: MS$_c$ improves BA-ECE from $0.043$ to $0.033$ and ACE from $0.069$ to $0.041$, while DC gives smaller gains.
These gains are unlikely to be explained solely by differences in model complexity, as MS and DC have the same number of trainable and identifiable parameters, while MS$_c$ has slightly fewer of both (Table~\ref{tab:num_params}). Rather, they further support the benefit of removing representation sensitivity.
On RoadSeg, all affine variants are close, consistent with the lower pooled free-energy variability reported in Section~\ref{sec:logit_offset_real}. The binary setting also makes MS$_c$ particularly restrictive: for $C=2$, MS$_c$ has only two identifiable parameters, compared with three for MS and DC (Table~\ref{tab:num_params}), which may explain why DC can be slightly preferable despite both variants being TI.
\rev{Paired intervals give the same picture. On Cityscapes, all calibration metrics are significantly improved by the TI counterparts of MS($\bar z$), except ACE, for which a paired confidence interval cannot be computed and significance cannot therefore be assessed. On RoadSeg, DC also significantly improves all calibration metrics, while the binary setting makes MS$_c$ too restrictive, as explained above. In particular, MS$_c$ is significantly worse than MS on BA-ECE (\appref{app:results}). Finally, on BraTS, the improvements from the TI counterparts are not significant.}

Taken together, these results support TI as a useful design principle for affine calibration: it makes calibration representation-consistent and improves or matches shift-sensitive counterparts. However, these affine calibrators are not constrained to preserve the argmax, so improving calibration can come at the cost of changing the predicted segmentation. Compared with the uncalibrated logit-pooled ensemble $\bar z$, the three calibration pipelines have negligible DSC changes on RoadSeg ($76.7$ to $76.5$--$76.7$), but large drops on BraTS ($76.9$ to $67.6$--$70.2$) and Cityscapes ($79.4$ to $70.4$--$73.4$). Thus, the calibration gains obtained by these affine methods are accompanied by substantial losses in segmentation accuracy on two of the three datasets. A flip audit for MS($\bar z$) shows that harmful flips generally outnumber beneficial ones: 50.4\% vs.\ 29.5\% on Cityscapes and 74.8\% vs.\ 18.6\% on BraTS, with the remaining flips leaving voxel-level accuracy unchanged. Visual examples of the spatial distribution of these flips are provided in \appref{app:flips}. This motivates the decision-preserving variants evaluated next in Point~3.

For LTS, we compare the original logit-based formulation LTS($\bar z$) from \cite{ding2021localtemperaturescaling}, which does not guarantee translation invariance, with our TI variant LTS($\log S(\bar z)$) (Table~\ref{tab:results}(c)).
This is a matched comparison: both variants use the same local temperature network, have the same number of parameters ($357$ on RoadSeg, $882$ on BraTS, and $3{,}507$ on Cityscapes; Table~\ref{tab:num_params}), and preserve the order by construction, so DSC is unchanged from the input segmentation.
Across the three datasets, the TI variant improves almost all calibration metrics, with BA-ECE on Cityscapes remaining unchanged. The gains are particularly clear on RoadSeg, where ECE decreases from $0.004$ to $0.002$ and ACE from $0.024$ to $0.007$, and on BraTS, where BA-ECE decreases from $0.134$ to $0.090$. \rev{However, only the ECE improvement on RoadSeg is significant.} LTS($\log S(\bar z)$) also improves ACE on BraTS ($0.237$ to $0.216$), although ACE remains worse than for the uncalibrated baselines. We attribute this to the severe foreground--background imbalance: the voxel-wise NLL objective and ECE are dominated by easy background voxels, whereas BA-ECE and ACE emphasize sparse uncertain regions and tumor boundaries. The flexible local temperature map can therefore fit the dominant confidence regime without reliably correcting calibration on rare foreground structures. 
Overall, using canonical inputs makes LTS representation-consistent and improves calibration without increasing model capacity, highlighting the practical benefit of translation invariance.

\paragraph{\textbf{3. Decision preservation exposes dataset-dependent calibration--segmentation effects (Table~\ref{tab:results}(d)).}}

We next isolate the effect of decision preservation within expressive affine calibration by comparing the unconstrained CDC pipeline with its argmax- and order-preserving counterparts, CMS$_{\mathrm{ap}}$ and CMS$_{\mathrm{op}}$. Unlike CDC, the CMS variants preserve the original segmentation map by construction and therefore retain the pooled DSC.
Across all datasets, unconstrained CDC induces decision changes. These changes are small on RoadSeg and BraTS, where CDC flips at most $0.6\%$ and $0.1\%$ of pixels, respectively, but become substantial on Cityscapes, with about $5\%$ of pixels changing label. Consequently, CDC consistently reduces DSC relative to the decision-preserving CMS variants, with the largest drops on Cityscapes ($69.3$ vs.\ $78.1$ for $\bar p$, and $74.2$ vs.\ $79.4$ for $\bar z$), followed by BraTS ($72.0$ vs.\ $76.0$ and $74.0$ vs.\ $76.9$). On RoadSeg, the effect is more modest, with DSC decreasing from $76.7$ to $75.7$.


The effect on calibration is both dataset- and metric-dependent. On Cityscapes, decision preservation does not reveal a clear calibration cost. While ECE is slightly increased or remains unchanged by the preservation constraints, both NLL and BA-ECE improve relative to CDC\rev{\ (the BA-ECE improvement is significant in all four comparisons and the NLL improvement in three of them)}. In particular, CMS${\mathrm{ap}}$ achieves the lowest NLL with both pooling strategies, whereas CMS${\mathrm{op}}$ yields the best BA-ECE. ACE also improves with order preservation, although the behavior of argmax preservation is less consistent across pooling strategies. Overall, on Cityscapes, preserving the decision map restores the pooled DSC while also improving most calibration metrics.
On BraTS, decision-preserving variants achieve very close NLL and ECE values compared with CDC, while improving BA-ECE\rev{\ (the BA-ECE improvement is significant in all four comparisons)}. ACE is more mixed and often slightly worse, especially for CMS$_{\mathrm{op}}$. Thus, preserving the decision map does not induce a clear degradation of calibration performance. 
RoadSeg is the only dataset where a clear calibration--segmentation trade-off is observed. CDC achieves the best calibration across all reported metrics\rev{, with significant improvements in most cases}. 
This may be partly explained by the binary nature of the problem. When C=2, the identifiable degrees of freedom decrease from 6 for CDC to only 4 for the decision-preserving CMS variants (Table~\ref{tab:num_params}), making the preservation constraints comparatively more restrictive.


Overall, enforcing argmax or order preservation prevents unintended segmentation changes by construction, but its calibration cost is not universal. In our experiments, a clear trade-off is observed only in the low-class RoadSeg setting, whereas on BraTS and Cityscapes the decision-preserving variants achieve comparable or even better calibration performance. This suggests that, when the conditional affine calibrator remains sufficiently expressive, decision-preserving constraints do not inherently compromise calibration\revdel{ and may even act as a beneficial regularizer}. Conversely, in binary or other low-class settings, the same constraints can reduce calibration flexibility and expose a more explicit calibration--segmentation trade-off.

\paragraph{\textbf{4. Practical implications for choosing a calibration pipeline.}}
When preservation of the deployed segmentation map is required by the  downstream task, decision-preserving constraints are mandatory. When this requirement is relaxed, these constraints \revdel{may still act as a beneficial regularizer for calibration}\rev{ reduce the effective capacity of the calibrator, with a calibration effect that varies by dataset}, and constrained and unconstrained calibrators should therefore be systematically compared\rev{, on a selection split held out from the data the calibrator is fitted on}. 
In contrast, translation invariance is both a structural requirement for representation-consistent post-hoc calibration and, in our matched comparisons, an empirically beneficial constraint, which we therefore recommend by default.

Among TI pipelines, methods can be categorized along two dimensions: calibration flexibility and decision preservation. On both Cityscapes and BraTS, no single method consistently outperforms all others across all calibration metrics. Nevertheless, simple temperature-based methods (TS and ETS), which are also order-preserving, achieve excellent performance on Cityscapes. On RoadSeg, LTS($\log S(\bar z)$), which provides a more flexible TI and order-preserving alternative to TS and ETS, yields the strongest ECE, BA-ECE, and ACE. On BraTS, CDC and the CMS variants perform particularly well. 
Overall, no single calibrator is uniformly optimal across datasets and evaluation metrics. \rev{Several calibrators should therefore be systematically compared on a selection split held out from the data used to fit them.} Reliability diagrams in \appref{app:reliability} provide additional qualitative support for these dataset-specific observations.

\color{black}




\section{Conclusion}

We showed that post-hoc calibration for semantic segmentation can be gauge-dependent: softmax logits are only identifiable up to an additive shift, and common calibrators can produce different outputs for equivalent representations of the same predictive distribution.
Motivated by this, we introduced representation-consistent calibration and derived a simple condition—translation invariance—that distinguishes well-defined calibrators (e.g., TS, DC) from representation-dependent ones (e.g., MS/VS and LTS).
We also presented practical ways to enforce translation invariance, including canonicalizing inputs via log-softmax and introducing new TI variants, such as constrained matrix scaling (MS$_c$).
\revdel{We further investigated the mismatch between training and calibration objectives and found that likelihood-driven calibration can change argmax decisions and degrade segmentation quality. To analyze this effect,}\rev{Since post-hoc calibration can degrade segmentation performance,} we introduced class-conditional affine calibrators and their argmax- and order-preserving counterparts, CMS$_{\mathrm{ap}}$ and CMS$_{\mathrm{op}}$, enabling matched comparisons that quantify the calibration--segmentation trade-off induced by decision preservation.
Across natural-image and medical segmentation benchmarks, including under covariate shift, translation invariance emerges as both a structural requirement and an empirically beneficial constraint in our matched comparisons.
Moreover, enforcing argmax or order preservation \rev{reduces the effective capacity of the calibrator, with a calibration effect that varies by dataset}\revdel{may induce an implicit regularization effect that can benefit calibration}.
Together, these results establish translation invariance and decision preservation as two fundamental design axes for post-hoc calibration in segmentation.

\subsection*{Broader Impact Statement}
By identifying translation invariance and decision preservation as structural properties of post-hoc calibration, this work provides practical design principles for future calibration methods in dense prediction. Translation invariance, in particular, is a lightweight, architecture-agnostic constraint that can be readily incorporated into a wide range of pipelines by using log probabilities instead of logits as input to the calibrator. Beyond its theoretical motivation, our results suggest that translation invariance is a beneficial design choice in practice and should therefore be considered by default when designing post-hoc calibration methods.

More generally, calibration plays a crucial role in safety-critical applications such as medical image segmentation by making predictive confidence better reflect observed accuracy, thereby improving downstream decision-making in both human and automated settings. It should therefore be viewed as a tool for decision support rather than a substitute for expert judgment.


\subsubsection*{Acknowledgments}
This work of the ITI HealthTech, as part of the ITI 2021-2028 program of the University of Strasbourg, CNRS and Inserm, was partially supported by IdEx Unistra (ANR-10-IDEX-0002) and SFRI (STRAT'US project, ANR-20-SFRI-0012) under the framework of the French Investments for the Future Program.
The authors would like to acknowledge the High Performance Computing Center of the University of Strasbourg for supporting this work by providing scientific support and access to computing resources.
Part of the computing resources were funded by the Equipex Equip@Meso project (Programme Investissements d'Avenir) and the CPER Alsacalcul/Big Data.

\bibliography{main}
\bibliographystyle{tmlr}

\clearpage
\appendix
\section*{Appendix}

This appendix contains the following items:
\begin{itemize}
    \item \hyperref[app:lts_ti]{\textbf{A. Additional Proofs}}
    \item \hyperref[app:hpo]{\textbf{B. Implementation and Training Details}}
    \item \hyperref[app:spatial]{\textbf{C. Spatial Calibration Assessment}}
    \item \hyperref[app:flips]{\textbf{D. Spatial Patterns of Label Flips}}
    \item \hyperref[app:ensemble_size]{\textbf{E. Sensitivity to Ensemble Size}}
    \item \hyperref[app:calib_size]{\textbf{F. Sensitivity to Calibration Set Size}}
    \item \hyperref[app:bin]{\textbf{G. Sensitivity to Binning}}
    \item \hyperref[app:ood]{\textbf{H. Calibration under Corruption-Based Covariate Shift}}
    \item \hyperref[app:results]{\textbf{I. Complete Quantitative Results with Uncertainty Intervals}}
    \item \hyperref[app:reliability]{\textbf{J. Reliability Diagrams}}
    \rev{\item \hyperref[app:robust]{\textbf{K. Robustness to the Segmentation Framework}}}
\end{itemize}

The calibration code is publicly released as the Fiducio library at \url{https://github.com/fiducio-ai/Fiducio}. It can be installed with \texttt{pip install fiducio}.

\clearpage

\section{Additional Proofs}
\subsection{Translation Invariance of Local Temperature Scaling}
\label{app:lts_ti}
We analyze when Local Temperature Scaling (LTS) is translation-invariant.
Let
\[
g_\theta^{\mathrm{LTS}}(\mathbf z(v))
=
S\!\left(
\frac{\mathbf z(v)}{\tau_\theta(v;\mathbf Z,\xi)}
\right),
\]
where $\tau_\theta(v;\mathbf Z,\xi)>0$ is a temperature predicted from the full logit field $\mathbf Z$ and optional features $\xi$.
Consider a spatially varying additive shift $c(v)$
from which we derive a full logit field $\mathbf Z'$ with $\mathbf z'(v)=\mathbf z(v)+c(v)\mathbf 1$.
If the temperature predictor is itself invariant to such shifts, we have: 
$\tau_\theta(v;\mathbf Z',\xi)=\tau_\theta(v;\mathbf Z,\xi)$.

For simplicity, and since this case is of limited practical relevance, we restrict ourselves to non-degenerate logits, i.e., $\mathbf{z}(v) \notin \mathrm{span}\{\mathbf{1}\}$. In the degenerate case, where all class logits are equal, the softmax output is uniform and independent of the temperature, so any violation of shift invariance of $\tau_\theta$ has no effect on the output of LTS.

Since softmax is invariant to additive constants, LTS is translation-invariant if the transformed logits differ only by an additive constant, i.e.,
\[
\frac{\mathbf z'(v)}{\tau_\theta(v;\mathbf Z',\xi)}
-
\frac{\mathbf z(v)}{\tau_\theta(v;\mathbf Z,\xi)}
\in \mathrm{span}\{\mathbf 1\}.
\]

If the temperature predictor is itself invariant to such shifts, we get: 
\[
\frac{\mathbf z'(v)}{\tau_\theta(v;\mathbf Z',\xi)}
-
\frac{\mathbf z(v)}{\tau_\theta(v;\mathbf Z,\xi)}
=
\frac{c(v)}{\tau_\theta(v;\mathbf Z,\xi)}\mathbf 1
\in \mathrm{span}\{\mathbf 1\}.
\]
Thus, LTS is translation-invariant.

Conversely, suppose that for some location $v$,
$\tau_\theta(v;\mathbf Z',\xi)\neq \tau_\theta(v;\mathbf Z,\xi)$.
Let
\[
\lambda'=\frac{1}{\tau_\theta(v;\mathbf Z',\xi)},
\qquad
\lambda=\frac{1}{\tau_\theta(v;\mathbf Z,\xi)}.
\]
Then
\begin{equation}
\label{eq:s}
\frac{\mathbf z'(v)}{\tau_\theta(v;\mathbf Z',\xi)}
-
\frac{\mathbf z(v)}{\tau_\theta(v;\mathbf Z,\xi)}
=
(\lambda'-\lambda)\mathbf z(v)+\lambda'c(v)\mathbf 1.
\end{equation}
Since $\mathbf z(v)\notin \mathrm{span}\{\mathbf 1\}$
and since $\lambda'\neq\lambda$, the vector of Eq. \ref{eq:s} is not in $\mathrm{span}\{\mathbf 1\}$.
Therefore, under the hypothesis that $\mathbf z(v)\notin \mathrm{span}\{\mathbf 1\} \forall v$, LTS is translation-invariant if and only if the temperature predictor is invariant to additive logit shifts. 

Note that the original LTS formulation, whose temperature network takes raw logits as input, does not guarantee translation invariance. In contrast, feeding the temperature network with canonical log-probabilities $\log S(\mathbf Z)$ makes $\tau_\theta$ invariant to additive shifts and yields a translation-invariant LTS pipeline.

\subsection{Order- and Argmax-Preserving Affine Calibrators Collapse to Temperature Scaling}
\label{app:collapse}

We first consider the bias-free case and write the calibrated probabilities as $g(\mathbf z)=S(\mathbf W\mathbf z)$, where $S$ is the softmax function.
For each class $i \in \{1,\dots,C\}$, we define the cone
$\Psi^{(i)}_1=\{\mathbf z \in \mathbb{R}^C \mid \mathbf z_i \ge \mathbf z_k,\ \forall k \neq i\}$,
i.e., the set of logits whose maximum is attained at class $i$.
A matrix $\mathbf W$ preserves the argmax if and only if each cone $\Psi^{(i)}_1$ is invariant under $\mathbf W$, i.e.,
$\mathbf z \in \Psi^{(i)}_1 \Rightarrow \mathbf W\mathbf z \in \Psi^{(i)}_1$.

As shown in Section~\ref{sec:apop}, the invariance of $\Psi^{(1)}_1$ (denoted $\Psi_1$ therein) under $\mathbf W$ is equivalent to the existence of a matrix $\mathbf W'$ such that:
$\mathbf W = \mathbf G^{-1}\mathbf W'\mathbf G$,
with
$\mathbf{W}'=\begin{pmatrix} a & \mathbf{r}^\top \\ \mathbf{0} & \mathbf{A} \end{pmatrix}$,
where $a \in \mathbb{R}$, $\mathbf{r} \in \mathbb{R}^{C-1}$, and $\mathbf{A} \in \mathbb{R}_+^{(C-1)\times(C-1)}$,
and $\mathbf G$ defined in Eq.~\ref{eq:G}. In the following, we denote by $\mathbf{W}^{(i)}$ a matrix leaving $\Psi^{(i)}_1$ invariant.
For simplicity, we illustrate the structure of $\mathbf{W}^{(1)}$ for $C=4$ and proceed with the proof without loss of generality in this setting.  A direct computation yields:
\[
\mathbf W^{(1)}=
\begin{pmatrix}
s & -r_1 & -r_2 & -r_3\\
s-\rho_2 & a_{11}-r_1 & a_{12}-r_2 & a_{13}-r_3\\
s-\rho_3 & a_{21}-r_1 & a_{22}-r_2 & a_{23}-r_3\\
s-\rho_4 & a_{31}-r_1 & a_{32}-r_2 & a_{33}-r_3
\end{pmatrix},
\quad
s=a+\sum_{i=1}^3 r_i,\quad
\rho_{i+1}=\sum_{j=1}^3 a_{ij}.
\]

The structure of $\mathbf W^{(1)}$ is fully determined by the following two constraints: (i) all rows have identical row sums; (ii) in each column $j>1$, the first entry is a minimal element. These constraints in turn imply that the first entry of the first column is a maximal element.

The characterization of $\mathbf W^{(i)}$ can be obtained from that of $\mathbf W^{(1)}$ by simultaneously permuting the $i$-th row and column with the first row and column. For instance, exchanging rows and columns $1$ and $2$ yields
\[
\mathbf{W}^{(2)} =
\begin{pmatrix}
a_{11}-r_1 & s-\rho_2 & a_{12}-r_2 & a_{13}-r_3\\
-r_1 & s & -r_2 & -r_3\\
a_{21}-r_1 & s-\rho_3 & a_{22}-r_2 & a_{23}-r_3\\
a_{31}-r_1 & s-\rho_4 & a_{32}-r_2 & a_{33}-r_3
\end{pmatrix}.
\]

By construction, $\Psi^{(2)}_1$ is invariant under $\mathbf W^{(2)}$. We now require consistency with the invariance of $\Psi^{(1)}_1$, i.e., that the ordering constraints derived for $\mathbf W^{(1)}$ remain valid under this permutation, while the row-sum condition is already preserved. We obtain in particular $a_{12}=a_{13}=0$, together with $\rho_2 \ge \rho_3$ and $\rho_2 \ge \rho_4$. The corresponding constraints on the first column, namely $a_{11} \ge a_{21}$ and $a_{11} \ge a_{31}$, become redundant once the constraints from $\Psi^{(3)}_1$ and $\Psi^{(4)}_1$ are incorporated, as these imply $a_{21}=a_{31}=0$.

Indeed, repeating the argument for $\Psi^{(3)}_1$ and $\Psi^{(4)}_1$ yields
$a_{21}=a_{23}=0$, $\rho_3 \ge \rho_2$, $\rho_3 \ge \rho_4$, and
$a_{31}=a_{32}=0$, $\rho_4 \ge \rho_2$, $\rho_4 \ge \rho_3$. Combining all these constraints, we conclude that $a_{ij}=0$ for all $i \neq j$, and $\rho_2 = \rho_3 = \rho_4 = \lambda > 0$. Finally, argmax-preserving matrices take the form
\begin{equation}
\label{eq:jenaimarre}
\mathbf W=
\begin{pmatrix}
s & -r_1 & -r_2 & -r_3\\
s-\lambda & \lambda-r_1 & -r_2 & -r_3\\
s-\lambda & -r_1 & \lambda-r_2 & -r_3\\
s-\lambda & -r_1 & -r_2 & \lambda-r_3
\end{pmatrix}.
\end{equation}

This can be expressed as $\mathbf W = \lambda \mathbf I + \mathbf 1 \mathbf c^\top$, where $\lambda > 0$ and $\mathbf I$ denotes the identity matrix. The vector $\mathbf c \in \mathbb{R}^C$ is defined by $c_1 = s-\lambda$ and $c_{i+1} = -r_i$ for $i \in \{1,2,3\}$.
In particular, $\mathbf c$ is arbitrary since both $s \in \mathbb{R}$ (corresponding to the free parameter $a$ in $\mathbf W'$) and $\mathbf r \in \mathbb{R}^{C-1}$ are unrestricted. In the general case ($C \geq 2$), the same reasoning applies within each of the $C$ cones.

We now consider the affine case $g(\mathbf z)=S(\mathbf W \mathbf z+\mathbf b)$. Since the argmax cones must remain invariant under translation, the bias vector must belong to all cones simultaneously, which implies $\mathbf b=k\mathbf 1$ ($k\in\mathbb R$).

Finally, using the translation invariance of the softmax, $S(\mathbf x + k \mathbf 1)=S(\mathbf x)$, we obtain
\[
S(\mathbf W \mathbf z + \mathbf b)
=
S(\lambda \mathbf z +  \mathbf 1 \mathbf c^\top \mathbf z + k \mathbf 1 )  
=
S(\lambda \mathbf z ) 
=
S\!\left(\frac{\mathbf z}{T}\right),
\qquad
T=\frac{1}{\lambda}.
\]

Therefore, MS calibrators that preserve the argmax
reduce exactly to TS. Since TS is also
order preserving, the same conclusion holds under 
order preservation.

\section{Implementation and Training Details}
\label{app:hpo}
\paragraph{Additional implementation details.}


All models are implemented in PyTorch~\citep{paszke2019pytorch}. The training loss is the voxel-wise cross-entropy (negative log-likelihood) for all methods, with the following Dirichlet-style regularizer described in~\cite{kull2019beyond} for affine calibrators (MS, MS$_c$, DC) :
\begin{equation}
\mathcal{R}(W,b)
=
\lambda \frac{1}{C(C-1)} \sum_{i\neq j} W_{ij}^2
+
\mu \frac{1}{C} \sum_{j=1}^{C} b_j^2,
\end{equation}
which penalizes off-diagonal weights and large biases. For CMS, the regularization is applied for each one of the $C$ models. For LTS, $\ell_2$ regularization is applied, following~\cite{ding2021localtemperaturescaling}.

Hyperparameters are optimized with Optuna~\citep{optuna_2019} using a TPE sampler and a budget of 20 trials per learned calibrator, selecting configurations by validation NLL on the validation split of each calibration repeat.
The search space is defined as follows:
\begin{itemize}
\item Learning rate (LR): log-uniform in $[10^{-4},\,3\times10^{-2}]$,
\item Early stopping patience (P): in $\{10,20,50\}$,
\item Regularization strengths $\lambda,\mu$: mixture of exact zero and log-uniform in $[10^{-6},\,10^{-1}]$,
\item Weight decay $\alpha$ for $\ell_2$ regularization (LTS only): mixture of exact zero and log-uniform in $[10^{-6},\,10^{-2}]$.
\end{itemize}

Optimization uses Adam, early stopping on validation NLL, and \texttt{ReduceLROnPlateau} scheduling. The $C$ class-wise experts of each class-conditional affine calibrator share the same hyperparameters.
All calibrators are parameterized to represent identity (absence of calibration) at initialization.

\paragraph{Local Temperature Scaling (LTS).}
Following \cite{ding2021localtemperaturescaling}, the temperature predictor $\tau_\theta$ follows a tree-style gated convolutional design, combining multiple candidate temperature maps with learned gates. Positivity is enforced with a softplus transform, and convolutions are applied in-plane with kernel size $5$ and dilation $2$.


\section{Spatial Calibration Assessment}
\label{app:spatial}

We compute the boundary-aware expected calibration error (BA-ECE) following~\cite{zeevi_spatially-aware_2025}. 
The main idea is to assess calibration in regions defined by their distance to the ground-truth boundary, so as to place more emphasis on locations where segmentation errors are typically more critical.

More precisely, let $d(v)$ denote the unsigned Euclidean distance (in voxels) from voxel $v$ to the ground-truth boundary, with $d(v)=0$ on the boundary itself. 
We partition the image into distance bands
\[
D_k = \{ v \mid d(v)\in \Delta_k \},
\]
using the default intervals:
$
\Delta_1=[0,3),\quad
\Delta_2=[3,7),\quad
\Delta_3=[7,15),\quad
\Delta_4=[15,+\infty).
$

BA-ECE is then computed over these regions as in~\cite{zeevi_spatially-aware_2025}, using inverse-distance weighting so that regions closer to the boundary contribute more strongly. Empty bands are ignored.

\section{Spatial Patterns of Label Flips}
\label{app:flips}

\begin{figure}[tb]
    \centering
    \includegraphics[width=\linewidth]{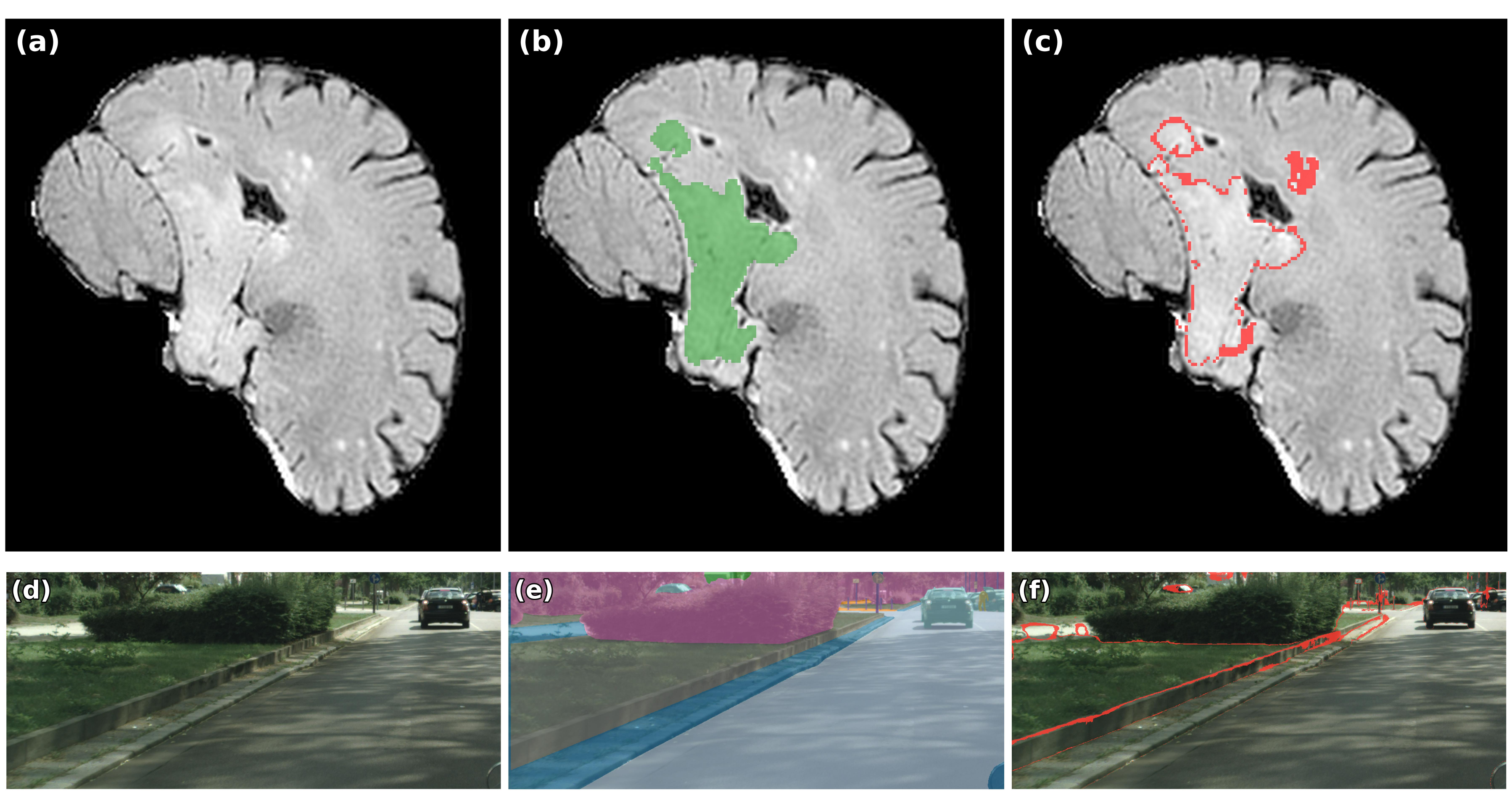}
\caption{Spatial patterns of label flips under $\mathrm{MS}(\bar{\mathbf z})$ on BraTS and Cityscapes. Top row: BraTS example with (a) MRI slice, (b) ground-truth foreground mask (green), and (c) flipped voxels after calibration (red). Bottom row: Cityscapes example with (d) input image, (e) ground-truth semantic segmentation overlay, and (f) flipped pixels after calibration (red). The Cityscapes view is cropped for visualization. In both cases, flips are concentrated near semantic boundaries, indicating that non--argmax-preserving calibration induces structured local label changes rather than spatially uniform perturbations.}
    \label{fig:brats_rsac_flip}
\end{figure}

Calibration may change the predicted class if it is not argmax-preserving. 
MS($\bar{z}$) is one such pipeline that may flip labels and alter the segmentation. 
Figure~\ref{fig:brats_rsac_flip} provides illustrative examples on BraTS and Cityscapes datasets, showing that label flips are not uniformly distributed: they concentrate primarily along tumor and semantic boundaries.
This occurs because likelihood-based calibration can locally alter the class ordering when margins are small, leading to spatially structured label changes.

\section{Sensitivity to Ensemble Size}
\label{app:ensemble_size}

Table~\ref{tab:offset_stats_ablation_fpooled} reports $\mathbb{E}[\mathrm{Std}_v(F_{\mathrm{pool}}(v))]$ for pooled ensembles of size $M\in\{1,2,3,4,5\}$. The statistic varies primarily by dataset (RoadSeg lowest, Cityscapes intermediate, BraTS highest) and
varies only slightly with $M$.
Importantly, substantial spatial variation is present for every tested ensemble size, including small ensembles ($M=1,2$), indicating that the offset variability motivating TI calibration is not an artifact of using $M=5$.

\begin{table}[t]
\centering
\small
\setlength{\tabcolsep}{6pt}
\renewcommand{\arraystretch}{.8}
\begin{tabular}{l c *{5}{c}}
\toprule
Dataset & {$n$} 
& {$M=1$} 
& {$M=2$} 
& {$M=3$} 
& {$M=4$} 
& {$M=5$} \\
\midrule
RoadSeg & 243
& 0.97 
& 1.04 
& 1.09 
& 1.06
& 1.08 \\
& 
& {\scriptsize [0.65, 0.96, 1.21]}
& {\scriptsize [0.70, 1.02, 1.30]}
& {\scriptsize [0.72, 1.08, 1.35]}
& {\scriptsize [0.71, 1.05, 1.31]}
& {\scriptsize [0.73, 1.06, 1.34]} \\
\addlinespace

BraTS & 271
& 8.41 
& 8.31 
& 8.99 
& 8.86
& 8.77 \\
& 
& {\scriptsize [8.21, 8.33, 8.52]}
& {\scriptsize [8.11, 8.23, 8.41]}
& {\scriptsize [8.78, 8.91, 9.09]}
& {\scriptsize [8.65, 8.78, 8.96]}
& {\scriptsize [8.56, 8.69, 8.87]} \\
\addlinespace

Cityscapes & 500
& 2.71 
& 2.67 
& 2.56 
& 2.56
& 2.54 \\
& 
& {\scriptsize [2.52, 2.66, 2.87]}
& {\scriptsize [2.46, 2.60, 2.83]}
& {\scriptsize [2.32, 2.47, 2.71]}
& {\scriptsize [2.32, 2.48, 2.73]}
& {\scriptsize [2.30, 2.46, 2.71]} \\


\bottomrule
\end{tabular}
\caption{\textbf{Spatial variability of pooled-logit offsets across ensemble sizes.} For each $M\in\{1,2,3,4,5\}$, we report $\mathbb{E}[\mathrm{Std}_v(F_{\mathrm{pool}}(v))]$. Main values are means across held-out cases; brackets show $[p_{25},p_{50},p_{75}]$.
Offset variability remains substantial for all ensemble sizes, motivating the use of translation-invariant calibration.}
\label{tab:offset_stats_ablation_fpooled}
\end{table}

\section{Sensitivity to Calibration Set Size}
\label{app:calib_size}

We vary the number of held-out samples used to fit the post-hoc calibrator:
$n \in \{5,\allowbreak 25,\allowbreak 50,\allowbreak 75,\allowbreak 99\}$
for Cityscapes and
$n \in \{3,\allowbreak 14,\allowbreak 27,\allowbreak 41,\allowbreak 50\}$
for BraTS. The test split is kept fixed throughout; increasing $n$ only changes
the calibration/validation partition of the non-test pool.
Performance is reported as the difference ($\Delta$) relative to
the reference configuration used in our main experiments, corresponding to
$n=50$ samples for both datasets.

For computational efficiency, we consider a representative subset of
calibrators covering different model capacities and constraints:
TS($\bar{z}$), MS($\bar{z}$), MS$_c$($\bar{z}$), LTS($\bar{z}$), and
CDC($\bar{z}$).

\begin{figure}[tb]
    \centering
    \includegraphics[width=\linewidth]{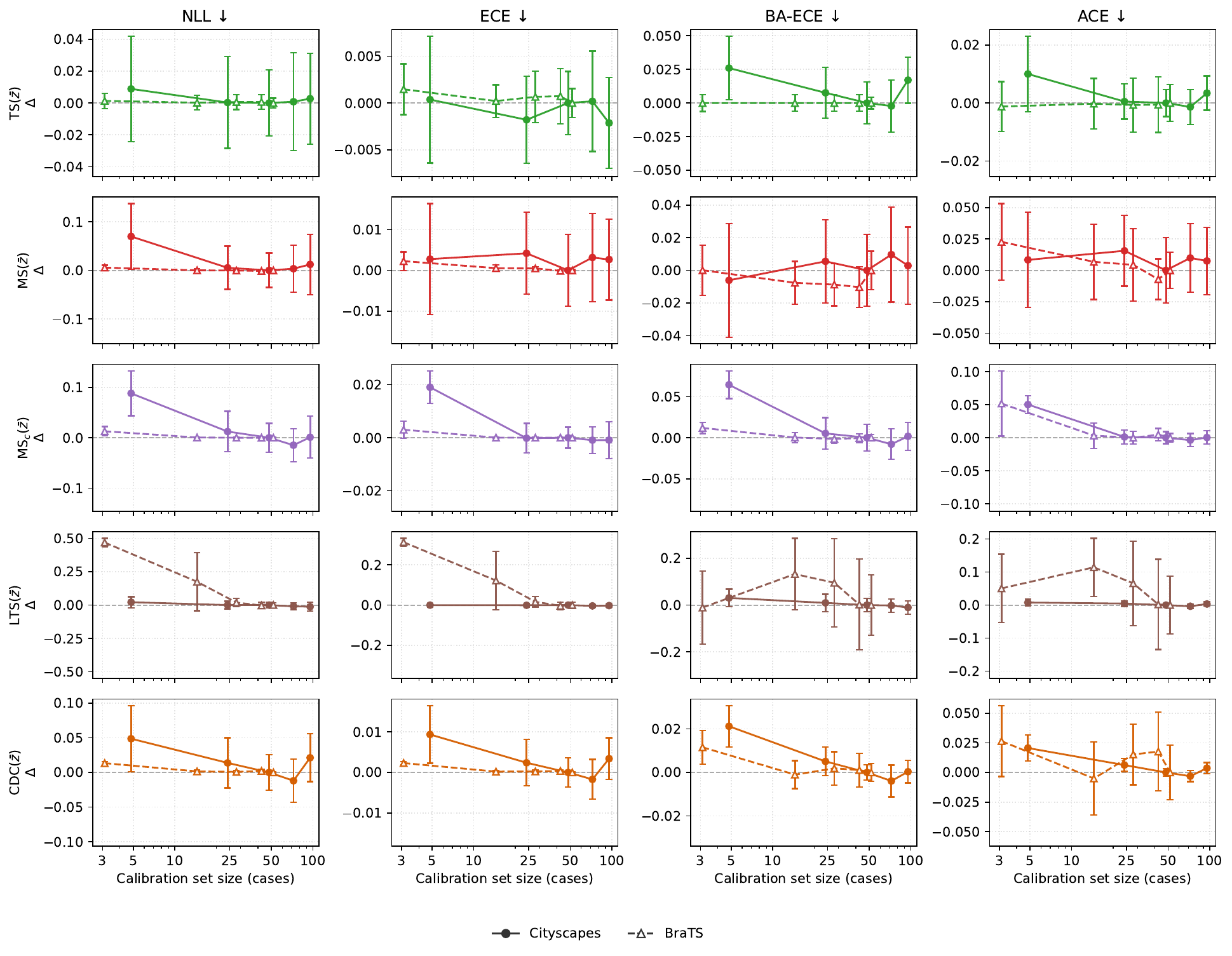}
    \caption{
    \textbf{Calibration set size sensitivity.}
    Effect of calibration set size $n$ on NLL, ECE, BA-ECE, and ACE for Cityscapes and BraTS, reported as $\Delta$ relative to the $n=50$ reference configuration used in the main experiments.
    }
    \label{fig:calib_size_ablation}
\end{figure}

Fig.~\ref{fig:calib_size_ablation} shows that performance is
generally stable around the reference budget of 50 calibration samples. The
largest variability occurs in the low-data regime, where the more expressive
calibrators can exhibit increased NLL or calibration error and wider confidence
intervals. Differences between moderate and larger calibration sets are small
relative to the uncertainty bands, supporting our use of $n=50$ as a pragmatic
trade-off between stable calibration and preserving held-out data.

\section{Sensitivity to Binning}
\label{app:bin}

\begin{figure}[tb]
    \centering
    \includegraphics[width=\linewidth]{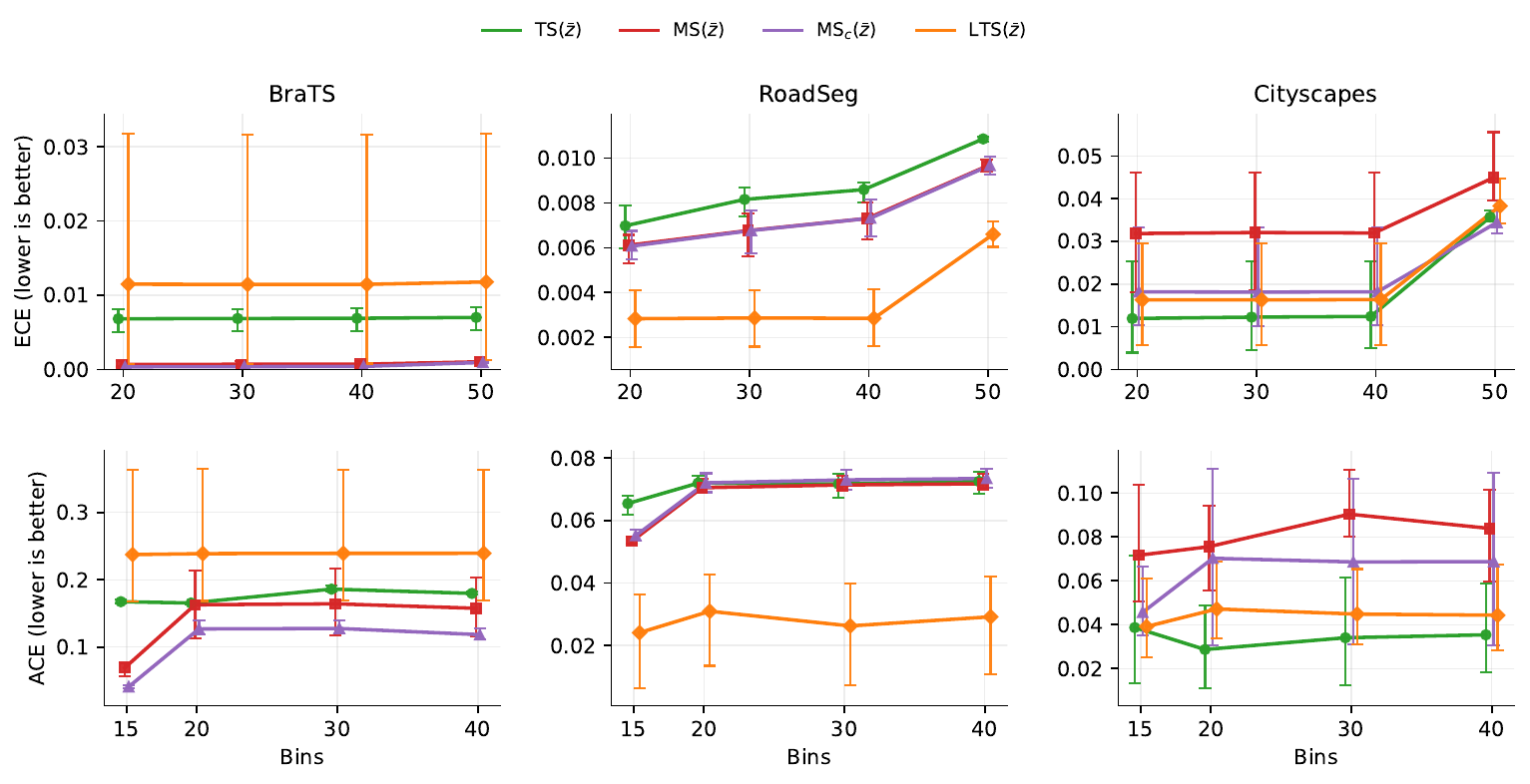}
    \caption{\textbf{Sensitivity of ECE and ACE to binning.} We report, for each dataset, ECE with $B\in\{20,30,40,50\}$ and ACE for $B\in\{15,20,30,40\}$.
    Error bars show 95\% bootstrap confidence intervals for ECE and [\textit{min}, \textit{max}] over repeats for ACE. 
    The ranking of methods is stable across bin counts, indicating that our conclusions are not driven by a particular choice of ECE/ACE discretization.}
    \label{fig:ece_ace_bins}
\end{figure}

Expected Calibration Error (ECE)~\citep{guo_calibration_2017} and ACE~\citep{kahl2024values} both depend on the discretization of the confidence axis.
To ensure our conclusions are not driven by a particular binning choice, we recompute ECE for $B\in\{20,30,40,50\}$ bins and ACE for $B\in\{15,20,30,40\}$ bins across datasets and methods (Fig.~\ref{fig:ece_ace_bins}).
Following \appref{app:calib_size}, we restrict this sensitivity analysis to a representative subset of calibrators.
Across datasets, changing the number of bins produces only small changes in ECE for most methods.
ACE is more sensitive in absolute value because it averages calibration gaps uniformly over non-empty bins; increasing the number of bins can therefore amplify sparsely populated confidence regions.
The range includes the two settings used in the main experiments: $B=15$ for ACE, chosen as a compromise between resolution, stability, and readable reliability diagrams, and $B=50$ for ECE.
The qualitative comparisons used in the paper do not change with binning.

\section{Calibration under Corruption-Based Covariate Shift}
\label{app:ood}


To assess robustness under covariate shift, we evaluate the Cityscapes calibrators from the main paper on corrupted test images. All calibrators are fitted on the clean in-distribution Cityscapes calibration set, exactly as in the main experiments, and are then evaluated without refitting on the corrupted test images. We focus on Cityscapes because standardized corruption protocols are available; RoadSeg shows limited sensitivity to calibration choices, and realistic covariate shifts for BraTS are beyond the scope of this work.

We generate two corrupted Cityscapes test sets with \emph{mild} and \emph{strong} severities, following the common-corruptions paradigm of ImageNet-C~\citep{hendrycks2019imagenetc} and its adaptation to Cityscapes~\citep{kamann2020benchmarking}. For each severity level, 
each image is transformed exactly once by sampling a perturbation type from a fixed perturbation bank using a deterministic hash of the filename and a fixed random seed. This yields a one-to-one correspondence between corrupted images and ground-truth labels and ensures reproducibility.
We restrict perturbations to photometric transformations in order to induce controlled covariate shift while preserving pixel-level label alignment. The perturbation bank consists of Gaussian noise, Gaussian blur, brightness scaling, JPEG compression, and fog/haze. No geometric transformations are applied (Table~\ref{tab:corruption_params}).

\begin{figure}[tb]
\centering
\includegraphics[width=\linewidth]{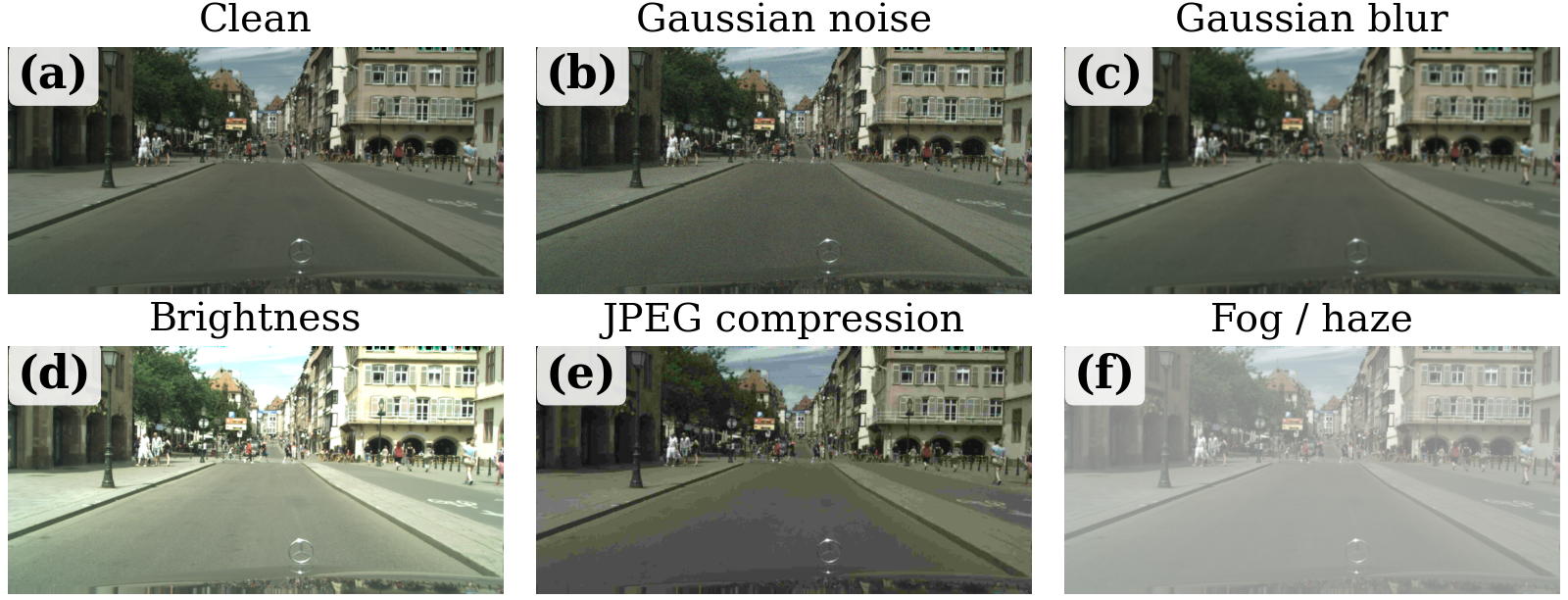}
\caption{\textbf{Cityscapes corruption types.} Example from the test set under \emph{strong} severity corruption: (\textbf{a}) in-distribution (clean) image; (\textbf{b}) Gaussian noise; (\textbf{c}) Gaussian blur; (\textbf{d}) brightness shift; (\textbf{e}) JPEG compression; (\textbf{f}) fog/haze. All variants preserve pixel-level correspondence with the original labels.}
\label{fig:cityscapes_ood_examples}
\end{figure}

Figure~\ref{fig:cityscapes_ood_examples} shows representative corrupted examples under \emph{strong} severity, and Table~\ref{tab:results_ood} reports the resulting calibration performance under corruption-based covariate shift. We discuss the main trends below.

\begin{table}[tb]
\centering
\small
\renewcommand{\arraystretch}{.85}
\setlength{\tabcolsep}{14pt}
\begin{tabular}{lcc}
\toprule
\textbf{Perturbation} & \textbf{Mild} & \textbf{Strong} \\
\midrule
Gaussian noise      & $\sigma=0.03$              & $\sigma=0.10$ \\
Gaussian blur       & $k=5,\ \sigma=1.0$         & $k=13,\ \sigma=3.0$ \\
Brightness shift    & factor $=1.2$              & factor $=1.8$ \\
JPEG compression    & quality $=60$              & quality $=10$ \\
Fog/haze          & $\alpha=0.1$               & $\alpha=0.5$ \\
\bottomrule
\end{tabular}
\caption{Photometric corruption settings used to induce test-time covariate shift on Cityscapes. For each corruption type, we define two severity levels, \textit{mild} and \textit{strong}. All corruptions are label-preserving and are applied only at evaluation time.}
\label{tab:corruption_params}
\end{table}

\providecommand{\NA}{\multicolumn{1}{c}{---}}
\providecommand{\ninv}{\textsuperscript{\(\times\)}}
\begin{table*}[tb]
\centering
\scriptsize
\setlength{\tabcolsep}{2pt}
\renewcommand{\arraystretch}{.86}
\resizebox{\textwidth}{!}{%
\begin{tabular}{@{}l
c c c c c c
c c c c c c
@{}}
\toprule
\multirow{2}{*}{\textbf{Pipeline}} &
\multicolumn{6}{c}{\textbf{Cityscapes  (\emph{mild} corruption)}} &
\multicolumn{6}{c}{\textbf{Cityscapes (\emph{strong} corruption)}} \\
\cmidrule(lr){2-7}\cmidrule(lr){8-13}
& {\textbf{NLL}$\downarrow$} & {\textbf{DSC}$\uparrow$} & {\textbf{ECE}$\downarrow$} & {\textbf{BA-ECE}$\downarrow$} & {\textbf{ACE}$\downarrow$} & {\textbf{Flip}$\downarrow$}
& {\textbf{NLL}$\downarrow$} & {\textbf{DSC}$\uparrow$} & {\textbf{ECE}$\downarrow$} & {\textbf{BA-ECE}$\downarrow$} & {\textbf{ACE}$\downarrow$} & {\textbf{Flip}$\downarrow$} \\
\midrule
\multicolumn{13}{@{}l}{\textbf{(a) Uncalibrated baselines}}\\
\addlinespace[1pt]
$p_0$ & 0.539& 70.2& 0.070& 0.195& 0.132& \NA& 1.173& 58.3& 0.153& 0.273& 0.179& \NA \\
$\bar{p}$ & 0.406& 77.2& 0.046& 0.142& 0.090& \NA& 0.756& 68.2& 0.087& 0.190& 0.112& \NA \\
$\bar{z}$ & 0.467& \bfseries 78.4& 0.062& 0.180& 0.128& \NA& 0.893& \bfseries 70.2& 0.124& 0.246& 0.164& \NA \\
\addlinespace[1pt]
\midrule
\multicolumn{13}{@{}l}{\textbf{(b) Translation invariance: MS vs.\ MS$_c$ and DC}}\\
\addlinespace[1pt]
\rowcolor{tirow} MS($\bar{z}$)\ninv & 0.445& 72.3& 0.048& 0.163& 0.088& 4.4& 0.949& 64.4& 0.110& 0.220& 0.129& 8.1 \\
\rowcolor{tirow} DC($\bar{z}$) & 0.352& 72.8& 0.037& 0.106& 0.071& \bfseries 2.7& 0.757& 63.6& 0.082& 0.160& 0.105& \bfseries 5.6 \\
\rowcolor{tirow} MS$_c$($\bar{z}$) & 0.359& 72.1& 0.037& 0.106& 0.070& 2.9& 0.764& 63.6& 0.090& 0.166& 0.108& 6.4 \\
\addlinespace[1pt]
\midrule
\multicolumn{13}{@{}l}{\textbf{(c) Translation invariance: LTS with logits vs.\ log-probabilities}}\\
\addlinespace[1pt]
\rowcolor{tirow} LTS($\bar{z}$)\ninv & 0.372& \bfseries 78.4& 0.040& 0.083& 0.074& \NA& 0.696& \bfseries 70.2& 0.072& 0.131& 0.095& \NA \\
\rowcolor{tirow} LTS($\log S(\bar{z})$) & 0.355&  \bfseries 78.4& 0.037& 0.077& 0.078& \NA& 0.656& \bfseries 70.2& 0.063& 0.115& 0.090& \NA \\
\addlinespace[1pt]
\midrule
\multicolumn{13}{@{}l}{\textbf{(d) Decision preservation: CDC vs CMS}}\\
\addlinespace[1pt]
\rowcolor{dprow} CDC($\bar{p}$) & 0.364& 68.2& \bfseries 0.035& 0.113& 0.073& 5.8& 0.730& 57.7& 0.076& 0.156& 0.099& 11.1 \\
\rowcolor{dprow} CMS$_{ap}$($\bar{p}$) & \bfseries 0.341& 77.2& 0.040& 0.091& 0.036& \NA& 0.702& 68.2& 0.070& 0.133& 0.061& \NA \\
\rowcolor{dprow} CMS$_{op}$($\bar{p}$) & 0.355& 77.2& 0.040& 0.075& \bfseries 0.030& \NA& 0.658& 68.2& 0.065& 0.113& \bfseries 0.038& \NA \\
\cmidrule(lr){2-13}
\rowcolor{dprow} CDC($\bar{z}$) & 0.369& 73.8& 0.037& 0.123& 0.078& 5.4& 0.773& 65.1& 0.085& 0.175& 0.114& 9.7 \\
\rowcolor{dprow} CMS$_{ap}$($\bar{z}$) & 0.354& \bfseries 78.4& 0.042& 0.099& 0.054& \NA& 0.742& \bfseries 70.2& 0.080& 0.153& 0.118& \NA \\
\rowcolor{dprow} CMS$_{op}$($\bar{z}$) & 0.368& \bfseries 78.4& 0.039& 0.086& 0.037& \NA& 0.698& \bfseries 70.2& 0.073& 0.138& 0.084& \NA \\
\addlinespace[1pt]
\midrule
\multicolumn{13}{@{}l}{\textbf{(e) Standard temperature-based calibrators (order-preserving, TI): TS/ETS}}\\
\addlinespace[1pt]
\rowcolor{poolrow} TS($\bar{p}$) & 0.356& 77.2& 0.039& 0.073& 0.078& \NA& \bfseries 0.653& 68.2& \bfseries 0.062& \bfseries 0.109& \bfseries 0.089& \NA \\
\rowcolor{poolrow} TS($\bar{z}$) & 0.368& \bfseries 78.4& 0.038& 0.086& 0.071& \NA& 0.686& \bfseries 70.2& 0.071& 0.136& 0.093& \NA \\
\cmidrule(lr){2-13}
\rowcolor{poolrow} ETS($\bar{p}$) & 0.356& 77.2& 0.038& 0.075& 0.077& \NA& \bfseries 0.653& 68.2& \bfseries 0.062& \bfseries 0.109& \bfseries 0.089& \NA \\
\rowcolor{poolrow} ETS($\bar{z}$) & 0.368& \bfseries 78.4& 0.038& 0.086& 0.071& \NA& 0.686& \bfseries 70.2& 0.071& 0.136& 0.093& \NA \\
\addlinespace[1pt]
\midrule
\multicolumn{13}{@{}l}{\textbf{(f) Standard affine calibrator (non order-preserving, TI): DC}}\\
\addlinespace[1pt]
\rowcolor{poolrow} DC($\bar{p}$) & 0.349& 66.6& 0.036& 0.103& \bfseries 0.069& 3.9& 0.736& 56.9& 0.079& 0.152& 0.098& 8.5 \\
\rowcolor{poolrow} DC($\bar{z}$) & 0.352& 72.8& 0.037& 0.106& 0.071& \bfseries 2.7& 0.757& 63.6& 0.082& 0.160& 0.105& \bfseries 5.6 \\
\addlinespace[1pt]
\bottomrule
\end{tabular}%
}
\caption{\textbf{Calibration on Cityscapes under covariate shift.} Numbered and lightly colored row blocks follow Table~\ref{tab:results}: blue for translation-invariance comparisons, green for decision-preservation comparisons, and beige for standard calibrators. DSC and Flip are reported in \%. Flip is the fraction of pixels whose predicted class changes relative to the corresponding uncalibrated pooled prediction; \textemdash\ indicates not applicable. \(\times\) denotes pipelines that are not translation-invariant. \textbf{Bold} indicates the best value per corruption severity and metric.}
\label{tab:results_ood}
\end{table*}

\paragraph{\textbf{1. Ensembles improve segmentation but do not reliably fix calibration.} (Table~\ref{tab:results_ood}(a))}
The same trends observed in-distribution persist under covariate shift. Probability pooling improves DSC over the single model (mild: 77.2 vs.\ 70.2; strong: 68.2 vs.\ 58.3), while logit pooling again achieves the highest segmentation accuracy (mild: 78.4; strong: 70.2). Pooling reduces some calibration errors under shift, especially for $\bar p$, but the effect remains pooling- and metric-dependent; ensembling alone therefore does not reliably correct miscalibration under covariate shift.

\paragraph{\textbf{2. TI pipelines appear more robust under shift.} (Table~\ref{tab:results_ood}(b))}
The in-distribution trends largely persist under covariate shift, but the gap between offset-sensitive and TI pipelines becomes more pronounced, especially for NLL and ECE. In the matched MS comparison, the unconstrained MS($\bar z$) consistently underperforms its two TI counterparts MS$_c$($\bar z$) and DC($\bar z$). 
A similar pattern is observed for LTS: using the TI input representation LTS($\log S(\bar z)$) improves NLL, ECE, and BA-ECE at both corruption severities. ACE is slightly worse under mild corruption (0.078 vs.\ 0.074), but better under strong corruption (0.090 vs.\ 0.095). Overall, these results suggest that TI pipelines are more robust under shift, with improvements in most comparisons, although the gains are not uniform across all metrics and severities.



\paragraph{\textbf 3. For Cityscapes, decision preservation improves segmentation without compromising calibration. (Table~\ref{tab:results_ood}(d))}
The pattern observed in-distribution for Cityscapes becomes more pronounced under strong shift: non-argmax-preserving pipelines can suffer large DSC drops. Under strong corruption, CDC($\bar p$) drops to 57.7 DSC with an 11.1\% flip rate, whereas CMS$_{\mathrm{ap}}$($\bar p$) preserves the pooled segmentation and maintains 68.2 DSC, with improvements in most calibration metrics.
A similar pattern holds for logit pooling. 
Overall, as in the in-distribution setting, 
enforcing decision preservation improves segmentation performance and also leads to improvements in most calibration metrics.

\section{Complete Quantitative Results with Uncertainty Intervals}
\label{app:results}

Table~\ref{tab:full_results} reports the quantitative results for all datasets, pipelines, and metrics, together with uncertainty intervals. The point estimates are identical to those reported in the main paper. Except for ACE, intervals are 95\% percentile confidence intervals (CI) computed using a hierarchical bootstrap over the $R=3$ repeats considered in the paper, resampling both repeats and cases within each repeat, following standard bootstrap methodology~\citep{efron1994introduction}. For ACE, the table reports the min--max range across repeat-level ACE values under the convention described in the evaluation protocol. The procedure was implemented using SciPy~\citep{virtanen2020scipy}.

\rev{Tables~\ref{tab:paired_ti} and~\ref{tab:paired_dp} report the paired difference for every matched comparison, computed as described in Section~\ref{sec:experiments}. The stars in Table~\ref{tab:results} and Table~\ref{tab:full_results} follow them.}

\providecommand{\NA}{\textemdash}
\providecommand{\ninv}{\,$\times$}

\begin{table*}[t]
\centering
\scriptsize
\setlength{\tabcolsep}{2pt}
\renewcommand{\arraystretch}{.85}
\begin{adjustbox}{max width=\textwidth,max totalheight=.88\textheight,center}
\begin{tabular}{@{}lcccccc@{}}
\toprule

\rowcolor{gray!20}
\multicolumn{7}{c}{\textbf{RoadSeg} \cite{mnih2013thesis}} \\
\cmidrule(lr){1-7}
\textbf{Pipeline} & \textbf{NLL}$\downarrow$ & \textbf{DSC (\%)}$\uparrow$ & \textbf{ECE}$\downarrow$ & \textbf{BA-ECE}$\downarrow$ & \textbf{ACE}$\downarrow$ & \textbf{Flip (\%)}$\downarrow$ \\
\midrule

 \multicolumn{7}{@{}l}{\textbf{(a) Uncalibrated baselines}}\\
\addlinespace[0.5pt]
$p_0$ & 0.091 [0.084, 0.099]& 76.3 [75.2, 77.4]& 0.017 [0.015, 0.018]& 0.094 [0.090, 0.098]& 0.119 [0.118, 0.119]& \NA \\
$\bar{p}$ & 0.085 [0.079, 0.092]& \bfseries 76.7 [75.6, 77.7]& 0.015 [0.014, 0.016]& 0.087 [0.084, 0.091]& 0.105 [0.105, 0.106]& \NA \\
$\bar{z}$ & 0.088 [0.083, 0.092]& \bfseries 76.7 [76.2, 77.3]& 0.016 [0.015, 0.017]& 0.091 [0.088, 0.093]& 0.117 [0.116, 0.117]& \NA \\
\addlinespace[0.5pt]
\midrule

\multicolumn{7}{@{}l}{\textbf{(b) Translation invariance: MS vs.\ MS$_c$ and DC}}\\
\addlinespace[0.5pt]
\rowcolor{tirow} MS($\bar{z}$)\ninv & 0.068 [0.063, 0.073]& 76.5 [75.4, 77.6]& 0.010 [0.009, 0.010]& 0.056 [0.052, 0.061]& 0.053 [0.053, 0.055]& 0.2 [0.2, 0.3] \\
\rowcolor{tirow} DC($\bar{z}$) & 0.067 [0.063, 0.072]\sig& \bfseries 76.7 [75.6, 77.7]\sig& 0.008 [0.008, 0.009]\sig& 0.047 [0.042, 0.052]\sig& 0.040 [0.038, 0.042]& \bfseries 0.1 [0.0, 0.1]  \\
\rowcolor{tirow} MS$_c$($\bar{z}$) & 0.068 [0.063, 0.073]& 76.5 [75.3, 77.5]\sig& 0.010 [0.009, 0.010]& 0.057 [0.052, 0.062]& 0.055 [0.053, 0.057]& 0.2 [0.2, 0.3]  \\
\addlinespace[0.5pt]
\midrule

\multicolumn{7}{@{}l}{\textbf{(c) Translation invariance: LTS with logits vs.\ log-probabilities}}\\
\addlinespace[0.5pt]
\rowcolor{tirow} LTS($\bar{z}$)\ninv & 0.067 [0.064, 0.070]& \bfseries 76.7 [76.2, 77.3]& 0.004 [0.003, 0.005]& 0.047 [0.038, 0.056]& 0.024 [0.006, 0.036]& \NA \\
\rowcolor{tirow} LTS($\log S(\bar{z})$) & 0.066 [0.063, 0.068]& \bfseries 76.7 [76.2, 77.3]& \bfseries 0.002 [0.001, 0.002]\sig& \bfseries 0.040 [0.040, 0.041]& \bfseries 0.007 [0.002, 0.009]& \NA  \\
\addlinespace[0.5pt]
\midrule

\multicolumn{7}{@{}l}{\textbf{(d) Decision preservation: CDC vs CMS}}\\
\addlinespace[0.5pt]
\rowcolor{dprow} CDC($\bar{p}$) & \bfseries 0.065 [0.060, 0.071]\sig& 75.7 [74.5, 76.8]& 0.006 [0.006, 0.007]\sig& 0.043 [0.037, 0.049]& 0.013 [0.012, 0.015]& 0.5 [0.5, 0.6]  \\
\rowcolor{dprow} CMS$_{ap}$($\bar{p}$) & 0.067 [0.062, 0.072]& \bfseries 76.7 [75.7, 77.7]\sig& 0.007 [0.006, 0.008]& 0.046 [0.041, 0.050]& 0.036 [0.034, 0.038]& \NA  \\
\rowcolor{dprow} CMS$_{op}$($\bar{p}$) & 0.067 [0.062, 0.072]& \bfseries 76.7 [75.7, 77.7]\sig& 0.007 [0.006, 0.008]& 0.046 [0.042, 0.051]& 0.037 [0.035, 0.039]& \NA  \\
\cmidrule(lr){2-7}
\rowcolor{dprow} CDC($\bar{z}$) & 0.066 [0.061, 0.071]\sig& 75.7 [74.6, 76.8]& 0.006 [0.006, 0.007]\sig& 0.041 [0.035, 0.045]\sig& 0.012 [0.009, 0.014]& 0.6 [0.5, 0.6]  \\
\rowcolor{dprow} CMS$_{ap}$($\bar{z}$) & 0.067 [0.062, 0.072]& \bfseries 76.7 [75.7, 77.9]\sig& 0.008 [0.007, 0.008]& 0.047 [0.042, 0.052]& 0.043 [0.040, 0.046]& \NA  \\
\rowcolor{dprow} CMS$_{op}$($\bar{z}$) & 0.067 [0.063, 0.072]& \bfseries 76.7 [75.6, 77.7]\sig& 0.007 [0.007, 0.008]& 0.047 [0.043, 0.052]& 0.043 [0.041, 0.044]& \NA  \\
\addlinespace[0.5pt]
\midrule

\multicolumn{7}{@{}l}{\textbf{(e) Standard temperature-based calibrators (order-preserving, TI): TS/ETS}}\\
\addlinespace[0.5pt]
\rowcolor{poolrow} TS($\bar{p}$) & 0.068 [0.065, 0.070]& \bfseries 76.7 [75.6, 77.7]& 0.008 [0.008, 0.008]& 0.059 [0.056, 0.063]& 0.056 [0.053, 0.058]& \NA \\
\rowcolor{poolrow} TS($\bar{z}$) & 0.068 [0.065, 0.071]& \bfseries 76.7 [76.2, 77.3]& 0.009 [0.008, 0.009]& 0.061 [0.057, 0.065]& 0.065 [0.062, 0.068]& \NA \\
\cmidrule(lr){2-7}
\rowcolor{poolrow} ETS($\bar{p}$) & 0.068 [0.065, 0.070]& \bfseries 76.7 [75.6, 77.7]& 0.008 [0.008, 0.009]& 0.059 [0.056, 0.063]& 0.055 [0.054, 0.058]& \NA \\
\rowcolor{poolrow} ETS($\bar{z}$) & 0.068 [0.066, 0.071]& \bfseries 76.7 [76.2, 77.3]& 0.009 [0.008, 0.009]& 0.061 [0.057, 0.065]& 0.065 [0.063, 0.068]& \NA \\
\addlinespace[0.5pt]
\midrule

\multicolumn{7}{@{}l}{\textbf{(f) Standard affine calibrator (non order-preserving, TI): DC}}\\
\addlinespace[0.5pt]
\rowcolor{poolrow} DC($\bar{p}$) & 0.067 [0.062, 0.072]& 76.6 [75.5, 77.7]& 0.008 [0.007, 0.008]& 0.046 [0.042, 0.052]& 0.033 [0.031, 0.034]& \bfseries 0.1 [0.1, 0.1] \\
\rowcolor{poolrow} DC($\bar{z}$) & 0.067 [0.063, 0.072]& \bfseries 76.7 [75.6, 77.7]& 0.008 [0.008, 0.009]& 0.047 [0.042, 0.052]& 0.040 [0.038, 0.042]& \bfseries 0.1 [0.0, 0.1] \\
\addlinespace[0.5pt]

\midrule
\rowcolor{gray!20}
\multicolumn{7}{c}{\textbf{BraTS 2024 GLI} \cite{brats2024}} \\
\cmidrule(lr){1-7}
\textbf{Pipeline} & \textbf{NLL}$\downarrow$ & \textbf{DSC (\%)}$\uparrow$ & \textbf{ECE}$\downarrow$ & \textbf{BA-ECE}$\downarrow$ & \textbf{ACE}$\downarrow$ & \textbf{Flip (\%)}$\downarrow$ \\
\midrule

\multicolumn{7}{@{}l}{\textbf{(a) Uncalibrated baselines}}\\
\addlinespace[0.5pt]
$p_0$ & 0.524 [0.520, 0.527]& 75.2 [73.6, 76.8]& 0.340 [0.338, 0.342]& 0.049 [0.045, 0.053]& 0.112 [0.110, 0.114]& \NA \\
$\bar{p}$ & 0.522 [0.518, 0.526]& 76.0 [74.5, 77.5]& 0.340 [0.337, 0.342]& 0.041 [0.037, 0.045]& 0.087 [0.084, 0.092]& \NA \\
$\bar{z}$ & 0.523 [0.522, 0.524]& \bfseries 76.9 [76.0, 77.4]& 0.340 [0.339, 0.341]& 0.046 [0.043, 0.048]& 0.096 [0.088, 0.100]& \NA \\
\addlinespace[0.5pt]
\midrule

\multicolumn{7}{@{}l}{\textbf{(b) Translation invariance: MS vs.\ MS$_c$ and DC}}\\
\addlinespace[0.5pt]
\rowcolor{tirow} MS($\bar{z}$)\ninv & 0.008 [0.006, 0.011]& 68.4 [63.0, 73.1]& 0.001 [0.001, 0.001]& 0.043 [0.032, 0.057]& 0.069 [0.057, 0.077]& \bfseries 0.1 [0.1, 0.2] \\
\rowcolor{tirow} DC($\bar{z}$) & 0.008 [0.007, 0.010]& 67.6 [57.2, 74.5]& 0.001 [0.001, 0.001]& 0.040 [0.032, 0.051]& 0.065 [0.038, 0.110]& \bfseries 0.1 [0.1, 0.1] \\
\rowcolor{tirow} MS$_c$($\bar{z}$) & \bfseries 0.007 [0.007, 0.008]& 70.2 [67.3, 72.6]& 0.001 [0.001, 0.001]& 0.033 [0.029, 0.037]& 0.041 [0.039, 0.043]& \bfseries 0.1 [0.1, 0.1] \\
\addlinespace[0.5pt]
\midrule

\multicolumn{7}{@{}l}{\textbf{(c) Translation invariance: LTS with logits vs.\ log-probabilities}}\\
\addlinespace[0.5pt]
\rowcolor{tirow} LTS($\bar{z}$)\ninv & 0.018 [0.005, 0.041]& \bfseries 76.9 [76.0, 77.4]& 0.012 [0.001, 0.032]& 0.134 [0.044, 0.301]& 0.237 [0.168, 0.363]& \NA \\
\rowcolor{tirow} LTS($\log S(\bar{z})$) & 0.014 [0.005, 0.028]&  \bfseries 76.9 [76.0, 77.4]& 0.008 [0.001, 0.021]& 0.090 [0.029, 0.189]& 0.216 [0.117, 0.304]& \NA \\
\addlinespace[0.5pt]
\midrule

\multicolumn{7}{@{}l}{\textbf{(d) Decision preservation: CDC vs CMS}}\\
\addlinespace[0.5pt]
\rowcolor{dprow} CDC($\bar{p}$) & 0.008 [0.007, 0.009]& 72.0 [68.6, 74.9]& 0.001 [0.001, 0.001]& 0.034 [0.030, 0.039]& 0.046 [0.039, 0.053]& \bfseries 0.1 [0.0, 0.1] \\
\rowcolor{dprow} CMS$_{ap}$($\bar{p}$) & \bfseries 0.007 [0.006, 0.008]\sig& 76.0 [74.5, 77.5]\sig& 0.001 [0.001, 0.001]& 0.029 [0.024, 0.034]\sig& 0.044 [0.031, 0.055]& \NA  \\
\rowcolor{dprow} CMS$_{op}$($\bar{p}$) & \bfseries 0.007 [0.006, 0.008]\sig& 76.0 [74.4, 77.5]\sig& 0.001 [0.001, 0.001]& \bfseries 0.028 [0.024, 0.033]\sig& 0.065 [0.052, 0.077]& \NA  \\
\cmidrule(lr){2-7}
\rowcolor{dprow} CDC($\bar{z}$) & 0.008 [0.006, 0.010]& 74.0 [72.6, 75.5]& 0.001 [0.001, 0.001]& 0.035 [0.027, 0.042]& \bfseries 0.033 [0.017, 0.057]& \bfseries 0.1 [0.0, 0.2] \\
\rowcolor{dprow} CMS$_{ap}$($\bar{z}$) & \bfseries 0.007 [0.007, 0.008]& \bfseries 76.9 [75.4, 78.4]\sig& 0.001 [0.001, 0.001]& 0.031 [0.026, 0.035]\sig& 0.041 [0.029, 0.057]& \NA  \\
\rowcolor{dprow} CMS$_{op}$($\bar{z}$) & \bfseries 0.007 [0.006, 0.008]& \bfseries 76.9 [75.3, 78.4]\sig& 0.001 [0.001, 0.001]& 0.030 [0.025, 0.035]\sig& 0.055 [0.039, 0.077]& \NA  \\
\addlinespace[0.5pt]
\midrule

\multicolumn{7}{@{}l}{\textbf{(e) Standard temperature-based calibrators (order-preserving, TI): TS/ETS}}\\
\addlinespace[0.5pt]
\rowcolor{poolrow} TS($\bar{p}$) & 0.025 [0.024, 0.025]& 76.0 [74.5, 77.5]& 0.005 [0.004, 0.006]& 0.046 [0.043, 0.048]& 0.146 [0.138, 0.157]& \NA \\
\rowcolor{poolrow} TS($\bar{z}$) & 0.029 [0.028, 0.029]& \bfseries 76.9 [76.0, 77.4]& 0.007 [0.005, 0.008]& 0.048 [0.045, 0.049]& 0.167 [0.165, 0.170]& \NA \\
\cmidrule(lr){2-7}
\rowcolor{poolrow} ETS($\bar{p}$) & 0.010 [0.010, 0.010]& 76.0 [74.5, 77.5]& \bfseries 0.000 [0.000, 0.000]& 0.047 [0.045, 0.049]& 0.166 [0.163, 0.169]& \NA \\
\rowcolor{poolrow} ETS($\bar{z}$) & 0.010 [0.010, 0.010]& \bfseries 76.9 [76.0, 77.4]& \bfseries 0.000 [0.000, 0.000]& 0.048 [0.046, 0.050]& 0.187 [0.182, 0.192]& \NA \\
\addlinespace[0.5pt]
\midrule

\multicolumn{7}{@{}l}{\textbf{(f) Standard affine calibrator (non order-preserving, TI): DC}}\\
\addlinespace[0.5pt]
\rowcolor{poolrow} DC($\bar{p}$) & 0.008 [0.007, 0.010]& 68.4 [63.2, 72.9]& 0.001 [0.001, 0.001]& 0.037 [0.030, 0.043]& 0.057 [0.029, 0.089]& \bfseries 0.1 [0.1, 0.1] \\
\rowcolor{poolrow} DC($\bar{z}$) & 0.008 [0.007, 0.010]& 67.6 [57.2, 74.5]& 0.001 [0.001, 0.001]& 0.040 [0.032, 0.051]& 0.065 [0.038, 0.110]& \bfseries 0.1 [0.1, 0.1] \\
\addlinespace[0.5pt]

\midrule
\rowcolor{gray!20}
\multicolumn{7}{c}{\textbf{Cityscapes} \cite{cordts2016cityscapes}} \\
\cmidrule(lr){1-7}
\textbf{Pipeline} & \textbf{NLL}$\downarrow$ & \textbf{DSC (\%)}$\uparrow$ & \textbf{ECE}$\downarrow$ & \textbf{BA-ECE}$\downarrow$ & \textbf{ACE}$\downarrow$ & \textbf{Flip (\%)}$\downarrow$ \\
\midrule

\multicolumn{7}{@{}l}{\textbf{(a) Uncalibrated baselines}}\\
\addlinespace[0.5pt]
$p_0$ & 0.485 [0.457, 0.515]& 71.7 [71.1, 72.3]& 0.063 [0.059, 0.067]& 0.182 [0.179, 0.185]& 0.125 [0.124, 0.126]& \NA \\
$\bar{p}$ & 0.377 [0.352, 0.401]& 78.1 [77.6, 78.6]& 0.044 [0.040, 0.048]& 0.136 [0.133, 0.139]& 0.079 [0.076, 0.083]& \NA \\
$\bar{z}$ & 0.434 [0.427, 0.448]& \bfseries 79.4 [79.3, 79.5]& 0.058 [0.057, 0.059]& 0.170 [0.168, 0.172]& 0.122 [0.120, 0.125]& \NA \\
\addlinespace[0.5pt]
\midrule

\multicolumn{7}{@{}l}{\textbf{(b) Translation invariance: MS vs.\ MS$_c$ and DC}}\\
\addlinespace[0.5pt]
\rowcolor{tirow} MS($\bar{z}$)\ninv & 0.404 [0.373, 0.439]& 73.4 [70.7, 75.9]& 0.045 [0.037, 0.055]& 0.156 [0.135, 0.179]& 0.072 [0.051, 0.104]& 3.8 [3.4, 4.2] \\
\rowcolor{tirow} DC($\bar{z}$) & 0.324 [0.301, 0.351]\sig& 70.4 [67.8, 73.5]& 0.035 [0.032, 0.038]\sig& 0.096 [0.083, 0.119]\sig& 0.040 [0.019, 0.070]& \bfseries 2.3 [2.1, 2.5]  \\
\rowcolor{tirow} MS$_c$($\bar{z}$) & 0.328 [0.304, 0.359]\sig& 72.8 [70.2, 76.6]& 0.034 [0.031, 0.038]\sig& 0.098 [0.085, 0.117]\sig& 0.046 [0.035, 0.066]& 2.4 [1.9, 2.8]  \\
\addlinespace[0.5pt]
\midrule

\multicolumn{7}{@{}l}{\textbf{(c) Translation invariance: LTS with logits vs.\ log-probabilities}}\\
\addlinespace[0.5pt]
\rowcolor{tirow} LTS($\bar{z}$)\ninv & 0.344 [0.339, 0.354]& \bfseries 79.4 [79.3, 79.5]& 0.016 [0.005, 0.029]& 0.075 [0.048, 0.106]& 0.039 [0.025, 0.061]& \NA \\
\rowcolor{tirow} LTS($\log S(\bar{z})$) & 0.342 [0.323, 0.366]& \bfseries 79.4 [79.3, 79.5]& \bfseries 0.010 [0.009, 0.010]& 0.075 [0.070, 0.080]& 0.029 [0.022, 0.042]& \NA  \\
\addlinespace[0.5pt]
\midrule

\multicolumn{7}{@{}l}{\textbf{(d) Decision preservation: CDC vs CMS}}\\
\addlinespace[0.5pt]
\rowcolor{dprow} CDC($\bar{p}$) & 0.335 [0.314, 0.361]& 69.3 [67.8, 71.2]& 0.032 [0.029, 0.036]& 0.108 [0.106, 0.111]& 0.052 [0.045, 0.058]& 5.1 [3.6, 6.0] \\
\rowcolor{dprow} CMS$_{ap}$($\bar{p}$) & \bfseries 0.314 [0.296, 0.338]\sig& 78.1 [77.6, 78.6]\sig& 0.035 [0.032, 0.039]& 0.086 [0.076, 0.104]\sig& 0.041 [0.030, 0.062]& \NA  \\
\rowcolor{dprow} CMS$_{op}$($\bar{p}$) & 0.326 [0.306, 0.350]\sig& 78.1 [77.6, 78.6]\sig& 0.035 [0.031, 0.038]& 0.071 [0.061, 0.086]\sig& 0.032 [0.020, 0.039]& \NA  \\
\cmidrule(lr){2-7}
\rowcolor{dprow} CDC($\bar{z}$) & 0.341 [0.319, 0.370]& 74.2 [73.6, 74.8]& 0.035 [0.031, 0.039]& 0.115 [0.110, 0.120]& 0.067 [0.062, 0.071]& 5.0 [4.4, 5.5] \\
\rowcolor{dprow} CMS$_{ap}$($\bar{z}$) & 0.327 [0.308, 0.353]\sig& \bfseries 79.4 [78.9, 79.8]\sig& 0.038 [0.033, 0.042]& 0.091 [0.082, 0.106]\sig& 0.081 [0.067, 0.095]& \NA  \\
\rowcolor{dprow} CMS$_{op}$($\bar{z}$) & 0.339 [0.317, 0.367]& \bfseries 79.4 [78.9, 79.8]\sig& 0.035 [0.031, 0.038]& 0.079 [0.069, 0.089]\sig& 0.041 [0.014, 0.056]& \NA  \\
\addlinespace[0.5pt]
\midrule

\multicolumn{7}{@{}l}{\textbf{(e) Standard temperature-based calibrators (order-preserving, TI): TS/ETS}}\\
\addlinespace[0.5pt]
\rowcolor{poolrow} TS($\bar{p}$) & 0.326 [0.320, 0.337]& 78.1 [77.6, 78.6]& 0.011 [0.007, 0.017]& \bfseries 0.069 [0.059, 0.086]& \bfseries 0.025 [0.020, 0.033]& \NA \\
\rowcolor{poolrow} TS($\bar{z}$) & 0.338 [0.332, 0.350]& \bfseries 79.4 [79.3, 79.5]& 0.012 [0.005, 0.025]& 0.079 [0.067, 0.099]& 0.039 [0.013, 0.071]& \NA \\
\cmidrule(lr){2-7}
\rowcolor{poolrow} ETS($\bar{p}$) & 0.326 [0.320, 0.337]& 78.1 [77.6, 78.6]& \bfseries 0.010 [0.005, 0.018]& 0.071 [0.060, 0.089]& 0.026 [0.019, 0.040]& \NA \\
\rowcolor{poolrow} ETS($\bar{z}$) & 0.338 [0.332, 0.350]& \bfseries 79.4 [79.3, 79.5]& 0.013 [0.005, 0.026]& 0.079 [0.066, 0.099]& 0.039 [0.012, 0.072]& \NA \\
\addlinespace[0.5pt]
\midrule

\multicolumn{7}{@{}l}{\textbf{(f) Standard affine calibrator (non order-preserving, TI): DC}}\\
\addlinespace[0.5pt]
\rowcolor{poolrow} DC($\bar{p}$) & 0.319 [0.292, 0.356]& 64.8 [62.0, 67.0]& 0.033 [0.030, 0.037]& 0.098 [0.079, 0.128]& 0.037 [0.020, 0.066]& 3.2 [2.8, 3.5] \\
\rowcolor{poolrow} DC($\bar{z}$) & 0.324 [0.301, 0.351]& 70.4 [67.8, 73.5]& 0.035 [0.032, 0.038]& 0.096 [0.083, 0.119]& 0.040 [0.019, 0.070]& \bfseries 2.3 [2.1, 2.5] \\
\addlinespace[0.5pt]

\bottomrule
\end{tabular}%
\end{adjustbox}
\caption{\small{\textbf{Complete calibration results on the test split.}
\textemdash\ indicates not applicable; $\times$ denotes non--translation-invariant pipelines.
\textbf{Bold} indicates the best mean value per dataset and metric\rev{\ and carries no claim of significance}. \rev{Each block pairs a base method with its counterparts. An $\ast$ on the base method indicates that it outperforms all of its counterparts; an $\ast$ on a counterpart indicates that it outperforms the base method. Here, ``outperforms'' means that the 95\% paired confidence interval for the metric difference excludes zero in favor of the indicated method (\appref{app:results}). ACE and Flip are not paired endpoints and therefore carry no stars.}}}
\label{tab:full_results}
\end{table*}

\clearpage
\begin{table}[H]
\centering\scriptsize
\setlength{\tabcolsep}{4pt}
\rev{\begin{tabular}{@{}lccc@{}}
\toprule
Paired difference & RoadSeg & BraTS & Cityscapes \\
\midrule
\addlinespace[2pt]
\multicolumn{4}{@{}l}{\textbf{Affine: MS($\bar z$) $\to$ MS$_c$($\bar z$)}}\\
\quad NLL & $-0.0000$ [$-0.0001$, $+0.0000$] & $-0.0009$ [$-0.0029$, $+0.0009$] & $-0.0763$ [$-0.1251$, $-0.0278$]$^{\ast}$ \\
\quad DSC & $+0.0003$ [$+0.0000$, $+0.0005$]$^{\ast}$ & $+0.0174$ [$-0.0186$, $+0.0443$] & $-0.0068$ [$-0.0350$, $+0.0103$] \\
\quad ECE & $+0.0000$ [$-0.0001$, $+0.0001$] & $-0.0001$ [$-0.0003$, $+0.0001$] & $-0.0105$ [$-0.0233$, $-0.0026$]$^{\ast}$ \\
\quad BA-ECE & $+0.0007$ [$+0.0002$, $+0.0012$]$^{\ast}$ & $-0.0099$ [$-0.0235$, $+0.0027$] & $-0.0576$ [$-0.0868$, $-0.0175$]$^{\ast}$ \\
\addlinespace[2pt]
\multicolumn{4}{@{}l}{\textbf{Spatial: LTS($\bar z$) $\to$ LTS($\log S(\bar z)$)}}\\
\quad NLL & $-0.0007$ [$-0.0016$, $+0.0007$] & $-0.0039$ [$-0.0360$, $+0.0212$] & $-0.0012$ [$-0.0059$, $+0.0034$] \\
\quad DSC & $+0.0000$ [$-0.0000$, $+0.0000$] & $+0.0000$ [$+0.0000$, $+0.0000$] & $+0.0000$ [$+0.0000$, $+0.0001$] \\
\quad ECE & $-0.0008$ [$-0.0013$, $-0.0002$]$^{\ast}$ & $-0.0035$ [$-0.0307$, $+0.0182$] & $-0.0032$ [$-0.0116$, $+0.0025$] \\
\quad BA-ECE & $-0.0016$ [$-0.0066$, $+0.0021$] & $-0.0437$ [$-0.2719$, $+0.1353$] & $+0.0112$ [$-0.0083$, $+0.0423$] \\
\addlinespace[2pt]
\multicolumn{4}{@{}l}{\textbf{Affine: MS($\bar z$) $\to$ DC($\bar z$)}}\\
\quad NLL & $-0.0004$ [$-0.0005$, $-0.0002$]$^{\ast}$ & $-0.0000$ [$-0.0026$, $+0.0016$] & $-0.0805$ [$-0.1274$, $-0.0354$]$^{\ast}$ \\
\quad DSC & $+0.0021$ [$+0.0016$, $+0.0026$]$^{\ast}$ & $-0.0084$ [$-0.0574$, $+0.0540$] & $-0.0311$ [$-0.0404$, $-0.0221$]$^{\ast}$ \\
\quad ECE & $-0.0013$ [$-0.0014$, $-0.0012$]$^{\ast}$ & $-0.0000$ [$-0.0004$, $+0.0003$] & $-0.0100$ [$-0.0214$, $-0.0031$]$^{\ast}$ \\
\quad BA-ECE & $-0.0088$ [$-0.0101$, $-0.0077$]$^{\ast}$ & $-0.0026$ [$-0.0189$, $+0.0101$] & $-0.0591$ [$-0.0926$, $-0.0151$]$^{\ast}$ \\
\bottomrule
\end{tabular}}
\revnewcaption{\textbf{Paired differences, translation invariance.} Mean difference, second method minus first, with its 95\% paired bootstrap interval. NLL, ECE and BA-ECE are lower-better; DSC is higher-better. $^{\ast}$ marks an interval excluding zero.}
\label{tab:paired_ti}
\end{table}

\begin{table}[H]
\centering\scriptsize
\setlength{\tabcolsep}{4pt}
\rev{\begin{tabular}{@{}lccc@{}}
\toprule
Paired difference & RoadSeg & BraTS & Cityscapes \\
\midrule
\addlinespace[2pt]
\multicolumn{4}{@{}l}{\textbf{CDC($\bar p$) $\to$ CMS$_{ap}$($\bar p$)}}\\
\quad NLL & $+0.0013$ [$+0.0010$, $+0.0016$]$^{\ast}$ & $-0.0009$ [$-0.0016$, $-0.0003$]$^{\ast}$ & $-0.0206$ [$-0.0244$, $-0.0174$]$^{\ast}$ \\
\quad DSC & $+0.0104$ [$+0.0086$, $+0.0125$]$^{\ast}$ & $+0.0400$ [$+0.0138$, $+0.0794$]$^{\ast}$ & $+0.0877$ [$+0.0692$, $+0.1004$]$^{\ast}$ \\
\quad ECE & $+0.0015$ [$+0.0012$, $+0.0017$]$^{\ast}$ & $+0.0001$ [$+0.0000$, $+0.0002$]$^{\ast}$ & $+0.0041$ [$+0.0004$, $+0.0070$]$^{\ast}$ \\
\quad BA-ECE & $+0.0026$ [$-0.0003$, $+0.0050$] & $-0.0047$ [$-0.0072$, $-0.0012$]$^{\ast}$ & $-0.0220$ [$-0.0323$, $-0.0030$]$^{\ast}$ \\
\addlinespace[2pt]
\multicolumn{4}{@{}l}{\textbf{CDC($\bar p$) $\to$ CMS$_{op}$($\bar p$)}}\\
\quad NLL & $+0.0013$ [$+0.0011$, $+0.0016$]$^{\ast}$ & $-0.0011$ [$-0.0018$, $-0.0006$]$^{\ast}$ & $-0.0083$ [$-0.0107$, $-0.0059$]$^{\ast}$ \\
\quad DSC & $+0.0104$ [$+0.0085$, $+0.0126$]$^{\ast}$ & $+0.0400$ [$+0.0137$, $+0.0796$]$^{\ast}$ & $+0.0876$ [$+0.0692$, $+0.1005$]$^{\ast}$ \\
\quad ECE & $+0.0016$ [$+0.0012$, $+0.0020$]$^{\ast}$ & $+0.0001$ [$-0.0000$, $+0.0002$] & $+0.0035$ [$-0.0010$, $+0.0067$] \\
\quad BA-ECE & $+0.0031$ [$-0.0001$, $+0.0054$] & $-0.0052$ [$-0.0078$, $-0.0018$]$^{\ast}$ & $-0.0378$ [$-0.0492$, $-0.0204$]$^{\ast}$ \\
\addlinespace[2pt]
\multicolumn{4}{@{}l}{\textbf{CDC($\bar z$) $\to$ CMS$_{ap}$($\bar z$)}}\\
\quad NLL & $+0.0017$ [$+0.0016$, $+0.0019$]$^{\ast}$ & $-0.0009$ [$-0.0025$, $+0.0003$] & $-0.0139$ [$-0.0186$, $-0.0101$]$^{\ast}$ \\
\quad DSC & $+0.0102$ [$+0.0082$, $+0.0128$]$^{\ast}$ & $+0.0287$ [$+0.0163$, $+0.0410$]$^{\ast}$ & $+0.0517$ [$+0.0473$, $+0.0562$]$^{\ast}$ \\
\quad ECE & $+0.0020$ [$+0.0018$, $+0.0023$]$^{\ast}$ & $+0.0002$ [$-0.0000$, $+0.0004$] & $+0.0042$ [$-0.0018$, $+0.0087$] \\
\quad BA-ECE & $+0.0063$ [$+0.0052$, $+0.0073$]$^{\ast}$ & $-0.0040$ [$-0.0097$, $-0.0000$]$^{\ast}$ & $-0.0242$ [$-0.0337$, $-0.0117$]$^{\ast}$ \\
\addlinespace[2pt]
\multicolumn{4}{@{}l}{\textbf{CDC($\bar z$) $\to$ CMS$_{op}$($\bar z$)}}\\
\quad NLL & $+0.0017$ [$+0.0015$, $+0.0019$]$^{\ast}$ & $-0.0011$ [$-0.0028$, $+0.0002$] & $-0.0016$ [$-0.0042$, $+0.0011$] \\
\quad DSC & $+0.0102$ [$+0.0082$, $+0.0129$]$^{\ast}$ & $+0.0287$ [$+0.0163$, $+0.0412$]$^{\ast}$ & $+0.0517$ [$+0.0473$, $+0.0563$]$^{\ast}$ \\
\quad ECE & $+0.0020$ [$+0.0017$, $+0.0023$]$^{\ast}$ & $+0.0001$ [$-0.0001$, $+0.0003$] & $+0.0016$ [$-0.0030$, $+0.0046$] \\
\quad BA-ECE & $+0.0063$ [$+0.0048$, $+0.0076$]$^{\ast}$ & $-0.0044$ [$-0.0099$, $-0.0003$]$^{\ast}$ & $-0.0366$ [$-0.0485$, $-0.0277$]$^{\ast}$ \\
\bottomrule
\end{tabular}}
\revnewcaption{\textbf{Paired differences, decision preservation.} Read as in Table~\ref{tab:paired_ti}.}
\label{tab:paired_dp}
\end{table}

\clearpage
\section{Reliability Diagrams}
\label{app:reliability}

\begin{figure}[H]
    \centering
    \includegraphics[width=0.96\textwidth]{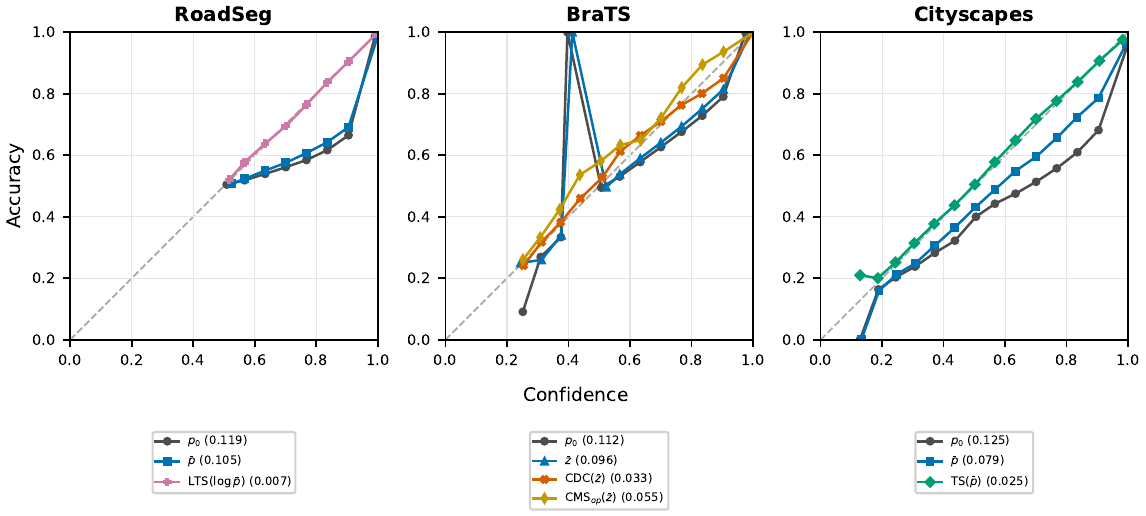}
    \caption{\textbf{Reliability diagrams.} Confidence--accuracy curves on the test split under the same ACE convention used in Table~\ref{tab:results}. Numbers in parentheses are mean test ACE across repeats. For each dataset, calibrated curves are selected by lowest ACE either over all calibrated methods or over the order-preserving subset; when both selections coincide, only one calibrated curve is shown. The dashed diagonal denotes perfect calibration.}
    \label{fig:ace15_reliability_appendix}
\end{figure}

Figure~\ref{fig:ace15_reliability_appendix} reports test-split reliability curves under the same convention as Table~\ref{tab:results}, with one panel per dataset comparing the single-model baseline, the uncalibrated pooling distribution, the best-ACE calibrated method selected without imposing order preservation, and the best-ACE order-preserving calibrated method.

\revnewsection{Robustness to the Segmentation Framework}
\label{app:robust}

\rev{All main results use nnU-Net ResEncM ensembles. This appendix drops the ensemble and changes the segmentation framework at the same time. We evaluate the official SegFormer-B0~\citep{xie2021segformer} Cityscapes checkpoint at a pinned revision against our own segmenter (with $M=1$), on the same 500 official validation images, unseen by both backbones, with the same calibration splits, test images, calibrator panel and metric code. Both arms are \emph{single} models, so no pooling is involved and the calibrators act on each model's own logits $z$. We follow Cityscapes' 19-class void-masked convention and restrict our $C=20$ nnU-Net logits accordingly. That restriction can itself change the nnU-Net predictions, so we compare within-model changes only, never absolute performance across backbones. The checkpoint also differs in pretraining and training protocol, so this is transportability evidence, not an architecture-controlled comparison.}

\rev{The pattern observed for Cityscapes in the core of the article (nnU-Net with $M=5$) also holds for $M=1$ and SegFormer-B0: decision preservation improves segmentation without compromising calibration, while enforcing translation invariance (TI) improves calibration in matched pipeline comparisons.}

\begin{table}[H]
\centering\footnotesize
\rev{\setlength{\tabcolsep}{2.8pt}\begin{tabular}{ll rrrr @{\hskip 1.0em} rrrr}
\toprule
& & \multicolumn{4}{c}{SegFormer-B0} & \multicolumn{4}{c}{nnU-Net ResEncM} \\
\cmidrule(lr){3-6}\cmidrule(lr){7-10}
Property tested & Matched difference & $\Delta$NLL $\downarrow$ & $\Delta$ECE $\downarrow$ & $\Delta$ACE $\downarrow$ & $\Delta$DSC $\uparrow$ & $\Delta$NLL $\downarrow$ & $\Delta$ECE $\downarrow$ & $\Delta$ACE $\downarrow$ & $\Delta$DSC $\uparrow$ \\
\midrule
Translation invariance & MS$_c$($z$)$-$MS($z$) & $-0.092$ & $-0.022$ & $-0.075$ & $+0.027$ & $-0.036$ & $-0.012$ & $-0.029$ & $+0.024$ \\
Decision preservation & CMS$_{\mathrm{ap}}$($z$)$-$CDC($z$) & $-0.001$ & $-0.002$ & $0.000$ & $+0.017$ & $-0.003$ & $-0.004$ & $-0.002$ & $+0.089$ \\
 & CMS$_{\mathrm{op}}$($z$)$-$CDC($z$) & $-0.003$ & $-0.002$ & $+0.009$ & $+0.017$ & $-0.004$ & $-0.007$ & $-0.021$ & $+0.089$ \\
\bottomrule
\end{tabular}}
\revnewcaption{\textbf{Robustness to the segmentation framework.} Mean matched difference, first method minus second, averaged over three fixed calibration splits.}
\label{tab:segformer}
\end{table}

\clearpage
\end{document}